\documentclass[conference]{IEEEtran}
\IEEEoverridecommandlockouts
\usepackage{cite}
\usepackage{amsmath,amssymb,amsfonts}
\usepackage{algorithmic}
\usepackage{graphicx}
\usepackage{textcomp}
\usepackage{xcolor}
\usepackage{algorithm}
\usepackage{algorithmic}
\usepackage{amsmath}
\usepackage{url}
\usepackage{longtable}
\usepackage{subcaption}
\usepackage{booktabs}
\usepackage{multirow}
\usepackage{threeparttable}
\usepackage[dvipsnames]{xcolor}
\usepackage{xspace}
\def\BibTeX{{\rm B\kern-.05em{\sc i\kern-.025em b}\kern-.08em
    T\kern-.1667em\lower.7ex\hbox{E}\kern-.125emX}}
\begin{document}

\title{
% {History Is Not Enough: Adaptive Financial Data Synthesis for Augmentation with a Curriculum Planner}
{History Is Not Enough: An Adaptive Dataflow System for Financial Time-Series Synthesis}
}

\author{
Haochong Xia$^{\textit{1,*}}$, Yao Long Teng$^{\textit{1,*}}$, Regan Tan$^{\textit{1}}$ , Molei Qin$^{\textit{1}}$ , Xinrun Wang$^{\textit{2}}$, Bo An$^{\textit{1,}\dagger}$\\[4pt]
\textit{$^{1}$College of Computing and Data Science, Nanyang Technological University, Singapore}\\
\textit{$^{2}$School of Computing and Information Systems, Singapore Management University, Singapore}\\[4pt]
\{HAOCHONG001, yaolong001, rtan072, molei001\}@e.ntu.edu.sg, xrwang@smu.edu.sg, boan@ntu.edu.sg
}

\maketitle

\begingroup
\renewcommand\thefootnote{}
\footnotetext{$^{*}$Equal contribution.}
\footnotetext{$^{\dagger}$Corresponding author.}
\endgroup

\maketitle

\begin{abstract}
% In quantitative finance, the discrepancy between training performance and real-world performance, particularly due to concept drift, presents a significant challenge. High accuracy on training data often fails to transfer to unseen data due to overfitting, undermining the models' value in real-world markets. Recognizing that historical data alone cannot fully capture market complexities, the mantra ``History Is Not Enough" underscores the need for augmented data generation. However, current data augmentation techniques do not adapt well to financial time series, and the workflow of applying augmented data in downstream financial tasks properly has not been thoroughly studied. In this paper, we propose a workflow for applying augmented data with an adaptive curriculum to address the uncertainty in downstream financial tasks. To synthesize diverse and high-quality financial data, we propose a data manipulation module involving single-stock transformation, multi-stock mix-up, and data curation techniques. The curriculum planner learns to control the manipulation of the training samples based on the state of the training data and the task model. Experimental results show that our plug-and-play workflow is model and task-agnostic, enhancing performance and reducing the risk of making poor decisions in dynamic market environments. Code is available at \href{https://anonymous.4open.science/r/HistoryIsNotEnough_/}{https://anonymous.4open.science/r/HistoryIsNotEnough\_/}

In quantitative finance, the gap between training and real-world performance—driven by concept drift and distributional non-stationarity—remains a critical obstacle for building reliable data-driven systems. Models trained on static historical data often overfit, resulting in poor generalization in dynamic markets. The mantra “History Is Not Enough” underscores the need for adaptive data generation that learns to evolve with the market rather than relying solely on past observations. We present a drift-aware dataflow system that integrates machine learning–based adaptive control into the data curation process. The system couples a parameterized data manipulation module comprising single-stock transformations, multi-stock mix-ups, and curation operations, with an adaptive planner–scheduler that employs gradient-based bi-level optimization to control the system. This design unifies data augmentation, curriculum learning, and data workflow management under a single differentiable framework, enabling provenance-aware replay and continuous data quality monitoring. Extensive experiments on forecasting and reinforcement learning trading tasks demonstrate that our framework enhances model robustness and improves risk-adjusted returns. The system provides a generalizable approach to adaptive data management and learning-guided workflow automation for financial data.

\end{abstract}


% \begin{figure*}[h!]
%     \centering
%     \includegraphics[width=0.75\textwidth]{figs/whole_pipeline_v4.pdf}
%     \caption{The workflow of training the planner and task model to learn a policy of controlling the data manipulation module with the validation loss of the task model. The training step of the planner is marked with (1), (2), (3), and $f_{\theta^{'}}$ is a copy of $f_{\theta}$. The fire icon marks the flow where parameters are updated.}
%     \label{fig:overall_workflow}
% \end{figure*}

% \input{contents/introduction}
\section{Introduction}

Machine learning techniques have been widely applied to quantitative finance research, encompassing tasks such as trading \cite{sun2024trademaster} and forecasting \cite{shahi2020stock}. The success of machine learning in quantitative finance largely stems from its ability to leverage the vast amount of financial data generated by markets \cite{sahu2023overview}. However, the constantly changing dynamics of the market present significant challenges. A fundamental assumption in applying machine learning techniques is that the data is independent and identically distributed (i.i.d.). When this i.i.d. assumption does not hold true, machine learning models tend to overfit the training data, which in turn reduces their robustness when applied to unseen data. The financial market, shaped by complex trader interactions, naturally evolves, leading to concept drift— inconsistency in the joint probability distribution between time 
$t$ and $t+k$ as $P_{t}(X,Y)\neq P_{t+k}(X,Y), k>0$, where $X$ is the feature and $Y$ is the target variable. For data-driven financial systems, such drift is not only a modeling challenge but also a data management problem: pipelines trained on static data lack mechanisms to adapt to distributional shifts over time. Addressing this gap requires an adaptive dataflow capable of managing evolving financial data streams.

% \textcolor{purple}{Still don't know what $P_t$ means. The joint probability of $(X,Y)$ at time $t$?}

Data manipulation techniques, such as data augmentation, play a crucial role in enhancing the robustness and generalization of machine learning models by expanding the training dataset to cover a broader range of potential market conditions \cite{shi2023neural}. Agent-based models \cite{samanidou2007agent} and deep generative models \cite{xia2024market} require significant computational resources and complex modeling. In contrast, data augmentation is simple and efficient. However, unlike in computer vision (CV), data augmentation is not yet extensively integrated into the quantitative finance pipeline, even though it can address both aleatoric and epistemic uncertainties \cite{hora1996aleatory, kapoor2022uncertainty}. One of the key challenges is the absence of a commonly agreed-upon benchmark for augmentation operations in time-series data, particularly in the context of financial data. This gap exists because augmentation can easily compromise the fidelity and correlations inherent in financial data. To address this, we design a parameterized data manipulation module that incorporates domain-specific constraints (e.g., K-line consistency, cointegration, and temporal non-stationarity) through single-stock transformations, curation steps, multi-stock mix-ups, and interpolation. This transforms augmentation from a heuristic preprocessing step into a controllable and auditable dataflow component that generates diverse yet faithful data.

% Given the diversity in the data, model and task in quantitative finance, developing effective augmentations for varying use cases require domain expertise and meticulous fine-tuning, making adapting implementation challenging. However, existing adaptive solvers for concept drift  are not designed to incorporate augmented data. For instance, approaches like DDG-DA \cite{li2022ddg} resample historical data but fail to address the epistemic uncertainty that arises when out-of-sample data is unseen during the training phase and has a strict assumption on the predictability of the concept drift. While there are adaptive augmentation curricula methods in CV domain, such as AdaAug \cite{cheung2021adaaug} and MADAug \cite{hou2023learn}. They only aim to learn representations from fixed datasets and are not designed to adapt to concept drift. In the absence of model- and task- agnostic workflows to address concept drift in financial data, we propose an adaptive planner which learns to provide optimal control of the parameter for the data manipulation module with the observation of the state of the data, the model and the training stage in the validation loop of the task model, and a scheduler to pace the augmentation proportion. 

Given the diversity in data, models, and tasks in quantitative finance, developing augmentations for varying use cases requires domain expertise and meticulous fine-tuning, making adaptation challenging. Existing adaptive approaches to data handling and augmentation \cite{li2022ddg, cheung2021adaaug, hou2023learn} address distributional bias in different ways: some rely solely on resampling existing data without generating new samples, while others employ augmentation strategies designed for fixed and stationary datasets. However, none of these methods continuously adapt their transformation policies as data and model states evolve. As a result, they are insufficient for non-stationary financial environments where concept drift is persistent and dynamic. From a data-system perspective, current pipelines lack a controller that can automatically regulate transformation operations based on feedback from downstream learning tasks. To fill this gap, we introduce a learning-guided planner that operates in the validation loop of the task model. This controller comprises an adaptive planner and a pacing scheduler which monitors both model state and data quality to determine optimal transformation probabilities and intensities. In effect, it serves as an autonomous workflow manager that dynamically adjusts data synthesis operations.

% Existing adaptive solvers are not designed to incorporate augmented data. For instance, approaches like DDG-DA \cite{li2022ddg} resample historical data but fail to address the epistemic uncertainty that arises when data is unseen during the training, relying on strict assumptions about the predictability of concept drift. While there are adaptive augmentation methods in the CV domain, such as AdaAug \cite{cheung2021adaaug} and MADAug \cite{hou2023learn}, they are primarily designed to learn representations from fixed datasets and are not equipped to adapt to concept drift. 

% Our goal is to enhance the current quantitative finance model training process by designing a plug-and-play learning strategy that can be seamlessly integrated into existing workflows. An adaptive augmentation method to address concept drift could be a promising solution. However, no such workflow exists in the financial domain.

% Hence, in the spirit of ``History is Not Enough", we propose a novel training workflow designed to address these challenges in quantitative finance. This workflow comprises a data manipulation module and an adaptive curriculum planner, enhancing generalization in quantitative finance machine learning tasks as shown in Fig.~\ref{fig:overall_workflow}. To the best of our knowledge, we are the first to introduce an adaptive augmentation workflow to quantitative finance machine learning tasks. Our major contributions include:

In the spirit of “History Is Not Enough,” we present a drift-aware adaptive dataflow system that unifies financial data synthesis, augmentation scheduling, and learning-based feedback control within a single workflow architecture. The system integrates the parameterized data manipulation module with a machine-learning-driven planner–scheduler that continuously optimizes transformation policies based on validation performance and overfitting signals. Together, these components constitute an adaptive data pipeline capable of improving data quality, diversity, and downstream model robustness under concept drift. To the best of our knowledge, this work provides the first learning-guided dataflow architecture tailored for financial time-series management. Our key contributions are summarized as follows:

\begin{enumerate}
\item \textbf{Adaptive dataflow framework.}
Motivated by empirical evidence of strong concept drift in financial data,
we design a unified \emph{adaptive dataflow system} that jointly \emph{generates}, \emph{curates}, and \emph{schedules} augmented data through a machine learning–based planner–scheduler architecture.

\item \textbf{Financially grounded synthesis module.}  
We develop a parameterized data manipulation module $\mathcal{M}$ that embeds financial priors into the transformation operations to ensure both realism and statistical diversity of the synthesized data.  

\item \textbf{Learning-guided augmentation control.}  
Our system employs a bi-level optimization scheme where an adaptive planner and an overfitting-aware scheduler dynamically coordinate manipulation strength and proportion of data to be manipulated according to model feedback. This enables self-adjusting data synthesis and curriculum pacing under concept drift, aligning data management decisions with learning dynamics.

\item \textbf{Empirical and systemic effectiveness.}  
Experiments on forecasting and reinforcement learning trading tasks show consistent gains in risk-adjusted performance while preserving data fidelity across various models and two mainstream financial markets.

\end{enumerate}

\begin{figure*}[h!]
    \centering
    \begin{subfigure}{0.135\textwidth}
        \centering
        \includegraphics[width=\linewidth]{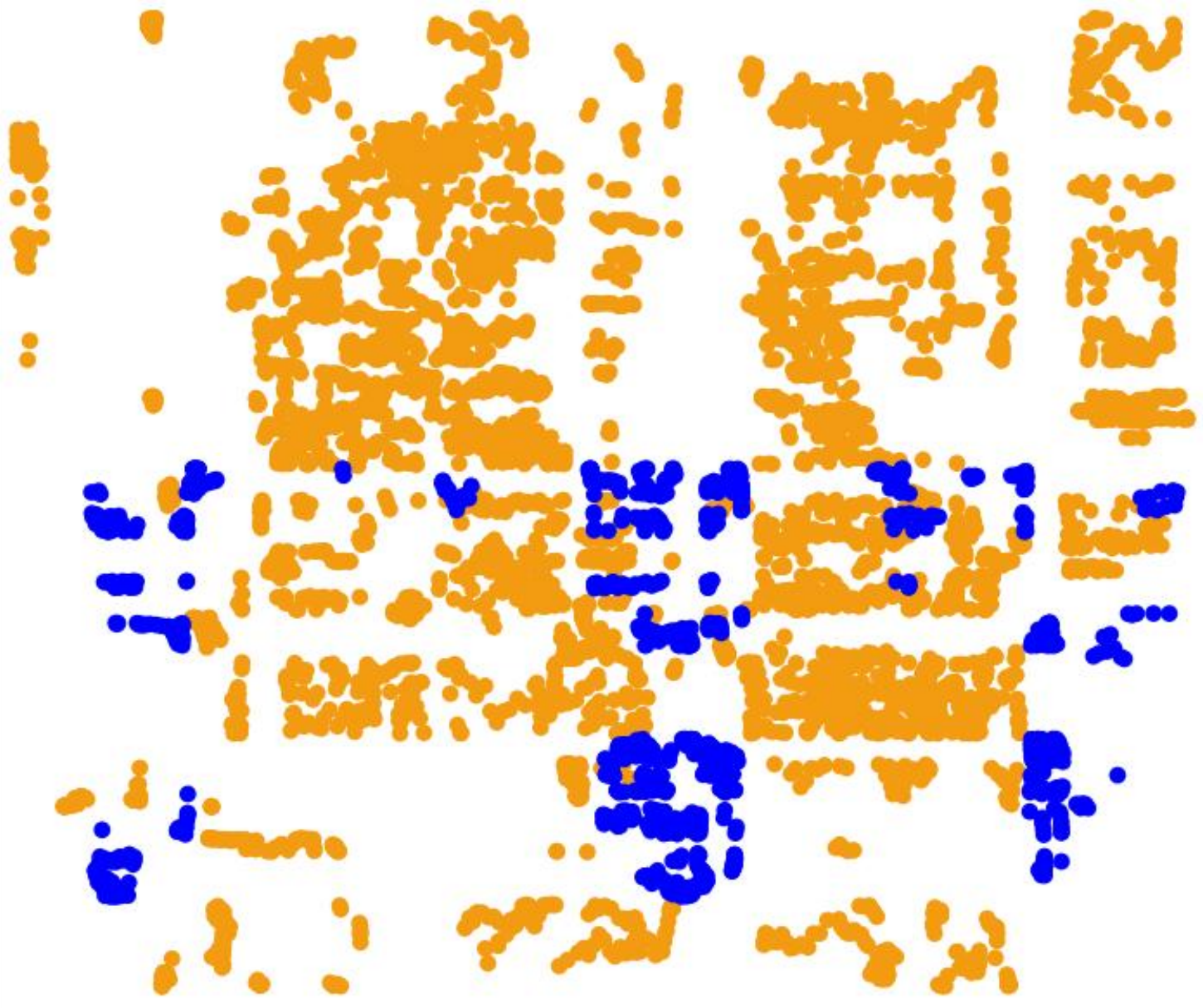}
        \captionsetup{font=small}
        \caption{MCD}
        \label{fig:sub1}
    \end{subfigure}%
    \hspace{0.02\textwidth}
    \begin{subfigure}{0.135\textwidth}
        \centering
        \includegraphics[width=\linewidth]{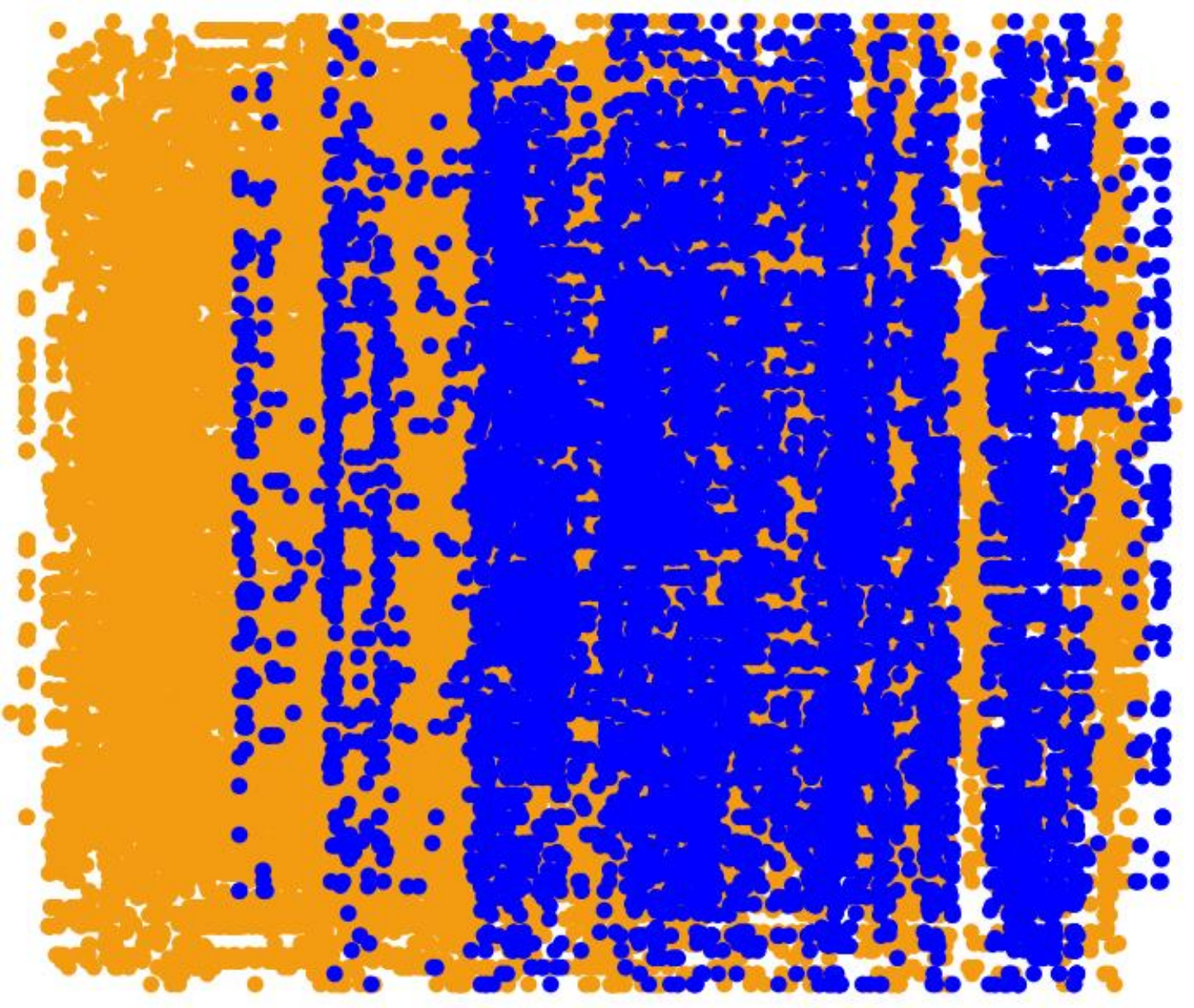}
        \captionsetup{font=small}
        \caption{Weather}
        \label{fig:sub2}
    \end{subfigure}%
    \hspace{0.02\textwidth}
    \begin{subfigure}{0.135\textwidth}
        \centering
        \includegraphics[width=\linewidth]{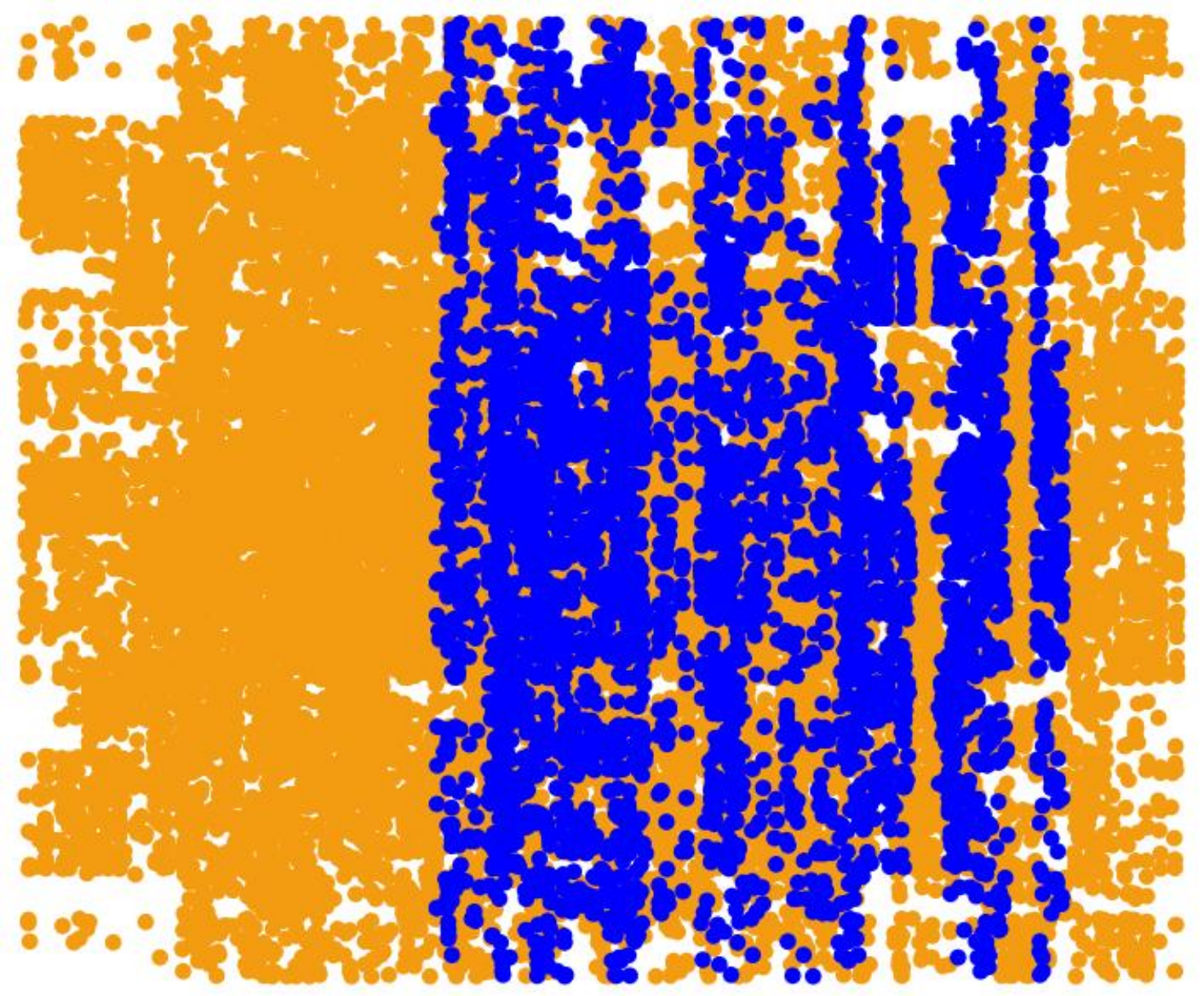}
        \captionsetup{font=small}
        \caption{Electricity}
        \label{fig:sub5}
    \end{subfigure}%
    \hspace{0.02\textwidth}
    \begin{subfigure}{0.135\textwidth}
        \centering
        \includegraphics[width=\linewidth]{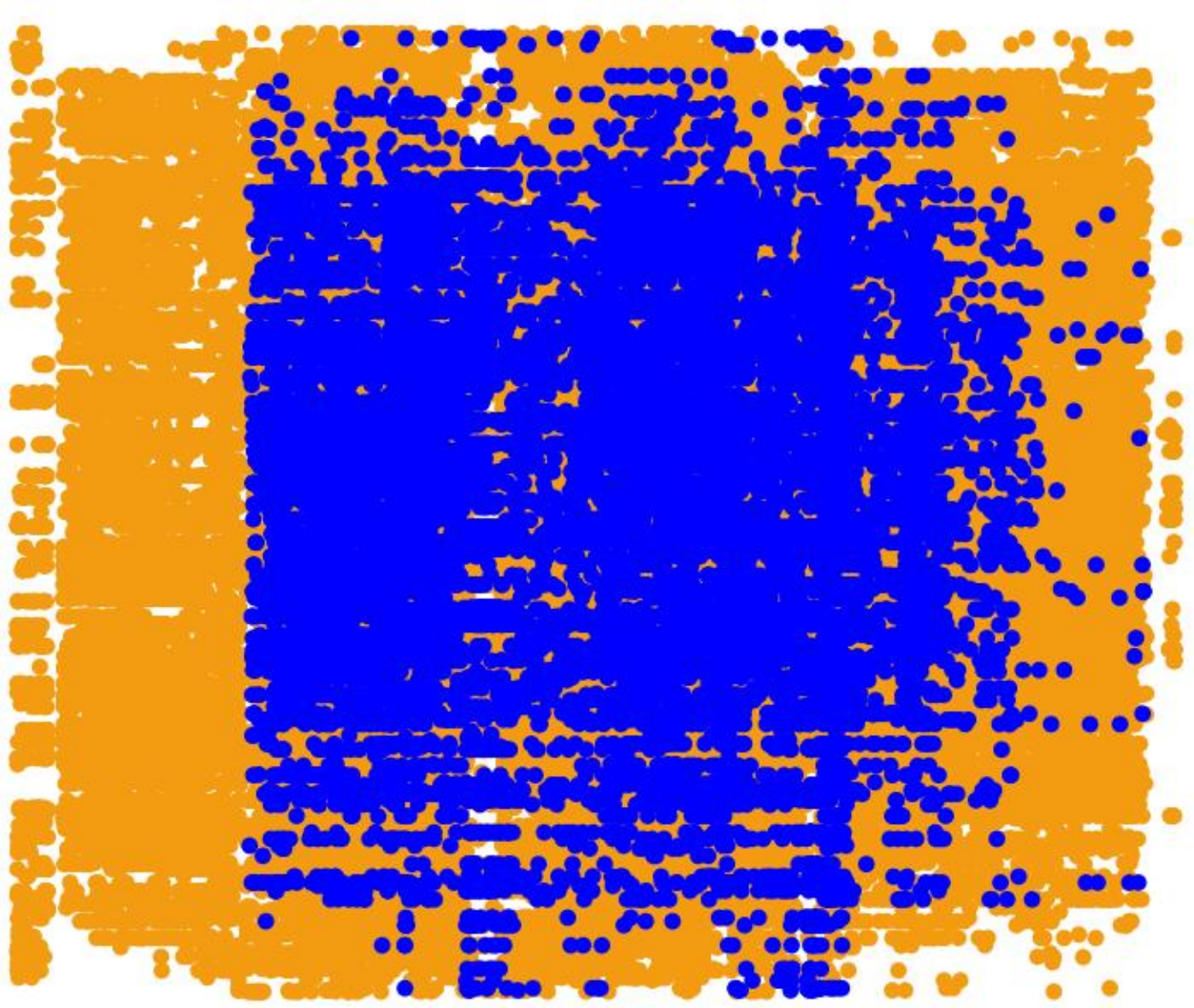}
        \captionsetup{font=small}
        \caption{ETT}
        \label{fig:sub3}
    \end{subfigure}%

    % Row 2
    \begin{subfigure}{0.135\textwidth}
        \centering
        \includegraphics[width=\linewidth]{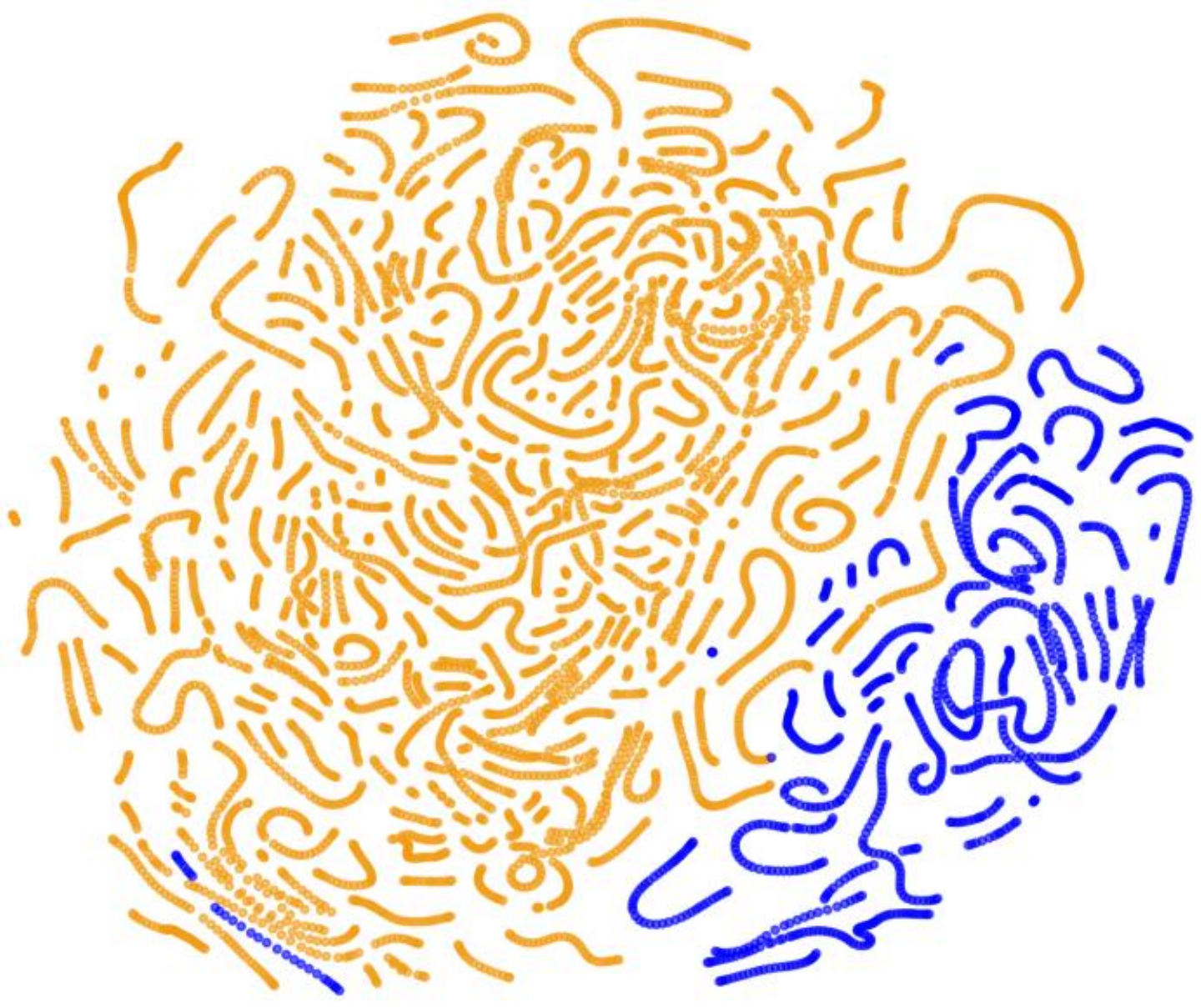}
        \captionsetup{font=small}
        \caption{MCD}
    \end{subfigure}%
    \hspace{0.02\textwidth}
    \begin{subfigure}{0.135\textwidth}
        \centering
        \includegraphics[width=\linewidth]{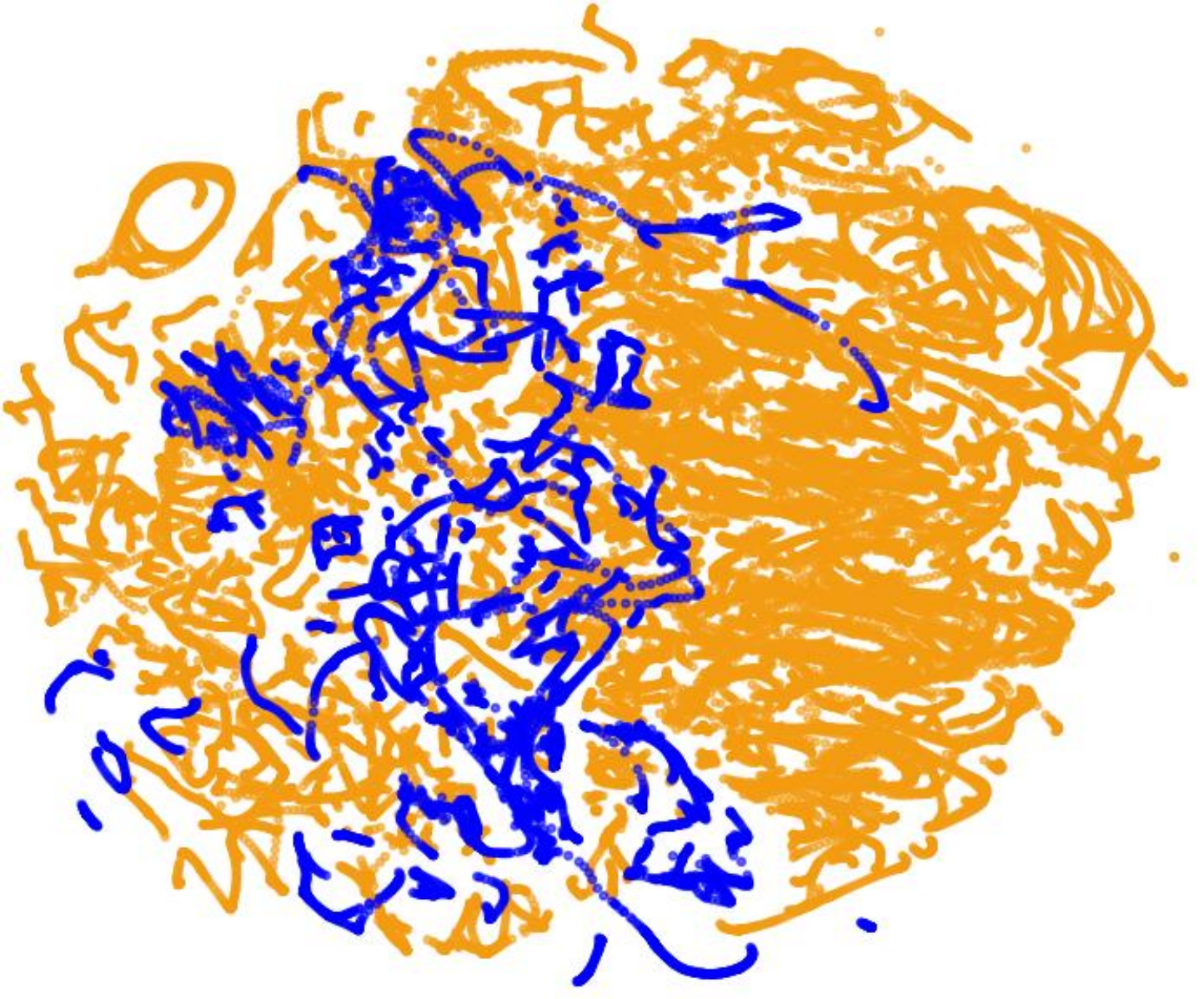}
        \captionsetup{font=small}
        \caption{Weather}
    \end{subfigure}%
    \hspace{0.02\textwidth}
    \begin{subfigure}{0.135\textwidth}
        \centering
        \includegraphics[width=\linewidth]{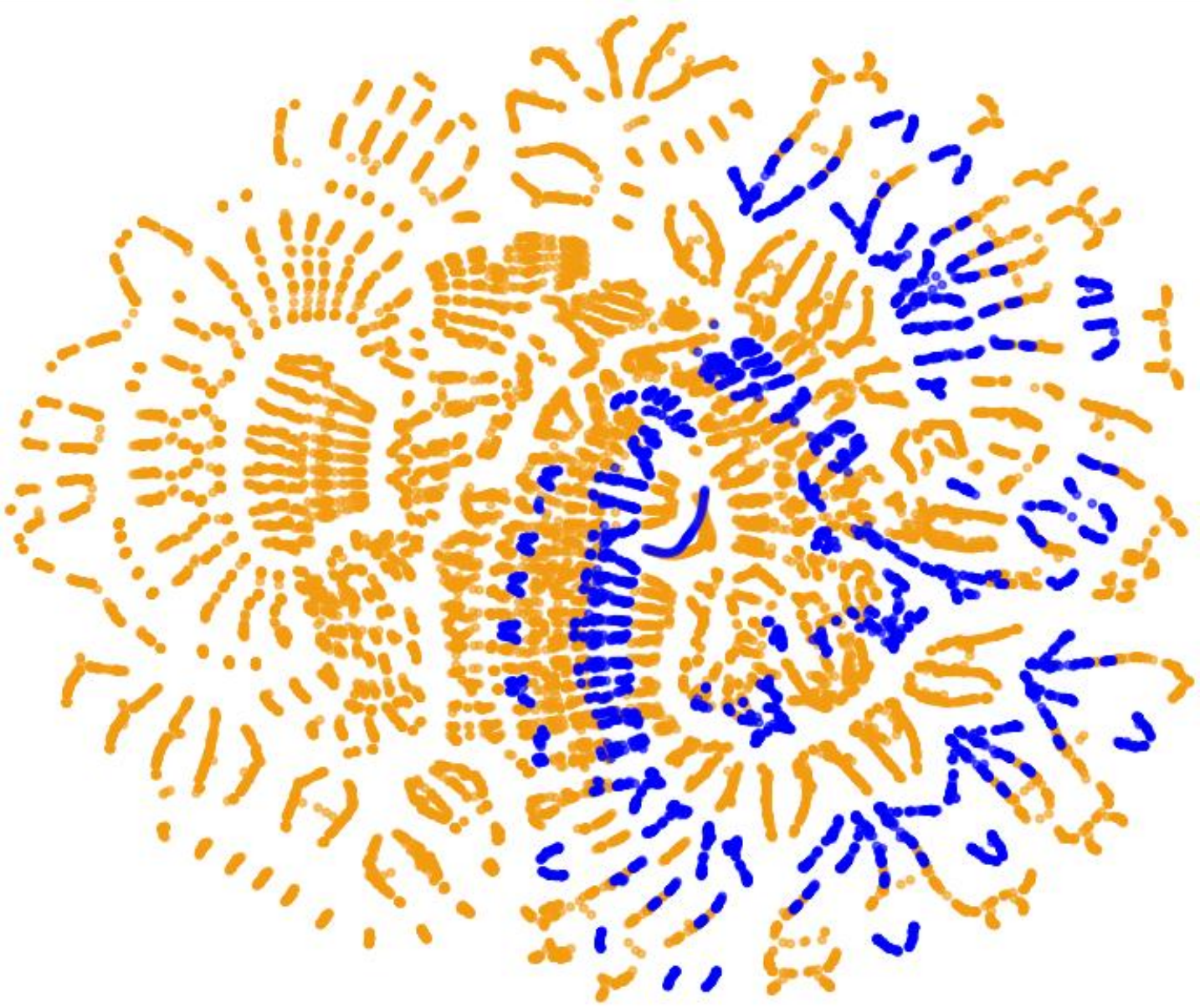}
        \captionsetup{font=small}
        \caption{Electricity}
    \end{subfigure}%
    \hspace{0.02\textwidth}
    \begin{subfigure}{0.135\textwidth}
        \centering
        \includegraphics[width=\linewidth]{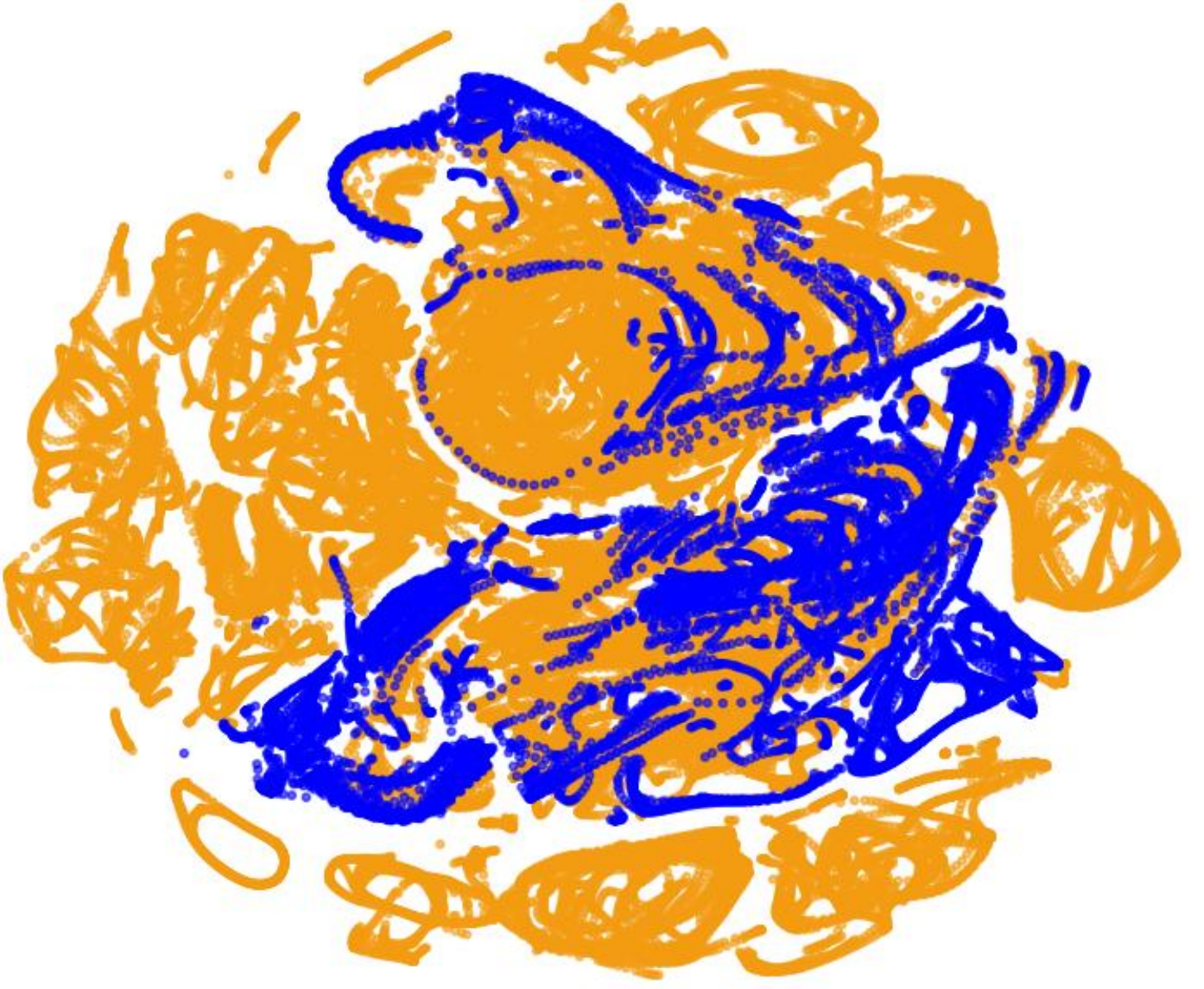}
        \captionsetup{font=small}
        \caption{ETT}
    \end{subfigure}%

    \caption{t-SNE visualizations to assess concept drift. Top row: t-SNE plot for 
$P(Y|X)$, where the x-axis represents the feature $X$ and the y-axis represents the daily return as the $Y$. Bottom row: t-SNE plot for $P(X)$. Orange dots mark the data in the training set, while blue dots mark the data in the test set.}
    \label{fig:T-SNE_combined}
\end{figure*}

\section{Background and Related Works}

Data augmentation techniques are important; however, they are less explored in time-series analysis. The method proposed by \cite{demirel2024finding} is specifically designed for non-stationary time-series. However, it requires the data to be periodic and is specifically designed for contrastive learning. There is still a lack of a benchmark for augmentation techniques for time-series, especially financial time-series data. Data augmentation techniques can be use-case specific, and finding the optimal policies requires domain knowledge and fine-tuning. Fixed policies often suffer from insufficient randomness. Conversely, automated augmentation like AutoAugment \cite{cubuk2018autoaugment} utilizes reinforcement learning to identify suitable augmentation policies, but requires extensive computational resources and restricts the randomness in its policies. 
% RandAugment \cite{cubuk2020randaugment} randomly selects the augmentation parameters.

Curriculum learning initiates training with simple patterns and progressively moves to more complex ones. Empirical evidence has shown that curriculum learning \cite{bengio2009curriculum} can enhance generalization. Recent works, such as AdaAug \cite{cheung2021adaaug} and MADAug \cite{hou2023learn}, have introduced the idea of curriculum learning accompanied by data augmentation techniques. AdaAug learns adaptive augmentation policies in a class and instance dependent manner. On top of this, MADAug applies a monotonic curriculum to introduce the augmented data to the training process. These methods are designed to learn robust image features. Consequently, there remains a lack of workflows that can generate effective augmentation policies tailored to the challenges of time-series data, particularly in the financial domain.

Generating augmented data has been an important way to expand historical data. Agent-based modeling \cite{samanidou2007agent,shi2023neural,zhang2024finagent} generates the data stream with the actions of autonomous agents. However, these models rely on the empirical behavior models of agents in the market which are subjective and computationally expensive, suffering from complex parameter tuning and thus lack generalizability \cite{gould2013limit, PhysRevE.76.016108,vyetrenko2020get}. Deep generative models generate data from an underlying distribution learned from historical data. TimeGAN \cite{yoon2019time} combines an autoencoder and GAN in a two-stage training to capture global and local temporal patterns. Diffusion-TS\cite{yuan2024diffusion} integrates seasonal-trend decomposition into a transformer-based diffusion model with a Fourier reconstruction loss, enabling interpretable multivariate time-series generation.
However, deep generative models need complex training and have less explainable evaluation metrics, complicating their integration into data management pipelines.

\section{Preliminaries}
\label{section:problem_formulation}

\subsection{Definitions}

\textbf{Financial Data and K-line Representation.}
Financial data are organized as sequences of \emph{K-line} (candlestick) tuples
\begin{equation}
    x_t = [O_t,\,H_t,\,L_t,\,C_t,\,V_t],
\end{equation}
where $O_t$, $H_t$, $L_t$, and $C_t$ denote the open, high, low, and close prices within an interval, and $V_t$ is the traded volume.
A valid K-line must satisfy the consistency constraint
\begin{equation}
    L_t \leq \min(O_t,C_t)\leq \max(O_t,C_t)\leq H_t,
\end{equation}
preserving market realism. These \emph{K-line features} encapsulate short-term momentum, volatility, and asymmetry of price movements.
They serve as the core components of both the forecasting input $\mathcal{X}$ and the RL state $s_t$.  
Our data-manipulation module (Section \ref{method}) explicitly enforces these constraints during augmentation to ensure economic plausibility.

\vspace{0.4em}
\noindent
\textbf{Time Series Forecasting.}
Let $\mathcal{X}=\{x_t\in\mathbb{R}^d\}_{t=1}^T$ denote a multivariate financial time series, where $x_t$ collects $d$ market features at timestamp $t$.
Given a look-back window of length $L$, a forecasting model
\begin{equation}
    f_\theta:\mathbb{R}^{d\times L}\rightarrow\mathbb{R}^{d'}
\end{equation}
predicts the next-period target $\hat{y}_t=f_\theta(x_{t-L+1:t})$.
The training objective minimizes the expected loss
\begin{equation}
    \min_{\theta}\ \mathbb{E}_{(x_t,y_t)\sim\mathcal{D}}
    [\,\ell(f_\theta(x_{t-L+1:t}),y_t)\,],
\end{equation}
where $\ell(\cdot)$ is the mean squared error (MSE).  

The prediction target $y_t$ is defined as the \emph{one-step close-to-close return}:
\begin{equation}
    y_t = \frac{C_{t+1} - C_t}{C_t},
\end{equation}
where $C_t$ and $C_{t-1}$ denote the closing prices at time $t$ and $t\!-\!1$, respectively.
This formulation captures short-term price dynamics and aligns with standard forecasting objectives in quantitative finance tasks.

\vspace{0.4em}
\noindent
\textbf{Reinforcement Learning (RL) for Trading.}
We formalize the trading environment as a Markov decision process (MDP)
\[
\mathcal{M}=(\mathcal{S},\mathcal{A},P,r,\gamma),
\]
where:
\begin{itemize}
    \item \emph{State space} $\mathcal{S}$:
    $s_t=[x_{t-L+1:t},\,p_t]$ concatenates recent features and the current position $p_t$.
    \item \emph{Action space} $\mathcal{A}$:
    discrete $\{-1,0,1\}$ for \emph{sell, hold, buy}.
    \item \emph{Transition probability function} $P(s_{t+1}|s_t,a_t)$:
    follows market evolution $x_{t+1}\!\sim\!P_X(\cdot|x_t)$.
    \item \emph{Reward function:}
    \begin{equation}
        r_t \;=\; p_{t-1}\, r^{\text{mkt}}_t \;-\; c\,|\Delta p_t|\,.
    \end{equation}
    where $r_t^{\text{mkt}}=\frac{C_{t+1}-C_t}{C_t}$ is the market return and $c$ the transaction-cost ratio.  
\end{itemize}
The agent $\pi_\theta(a_t|s_t)$ maximizes
\begin{equation}
    J(\pi_\theta)=\mathbb{E}_{\pi_\theta}\big[\sum_{t=0}^{T-1}\gamma^t r_t\big],
\end{equation}
with value function
\begin{equation}
    Q^*(s,a)=r(s,a)+\gamma\,\mathbb{E}_{s'\!\sim\!P}[\,\max_{a'}Q^*(s',a')\,],
\end{equation}
and optimal policy $\pi^*(s)=\arg\max_a Q^*(s,a)$.
This setting aligns with the RL trading experiments in Table \ref{tab:RL_result}.

\begin{figure*}[h!]
    \centering
    \includegraphics[width=\textwidth]{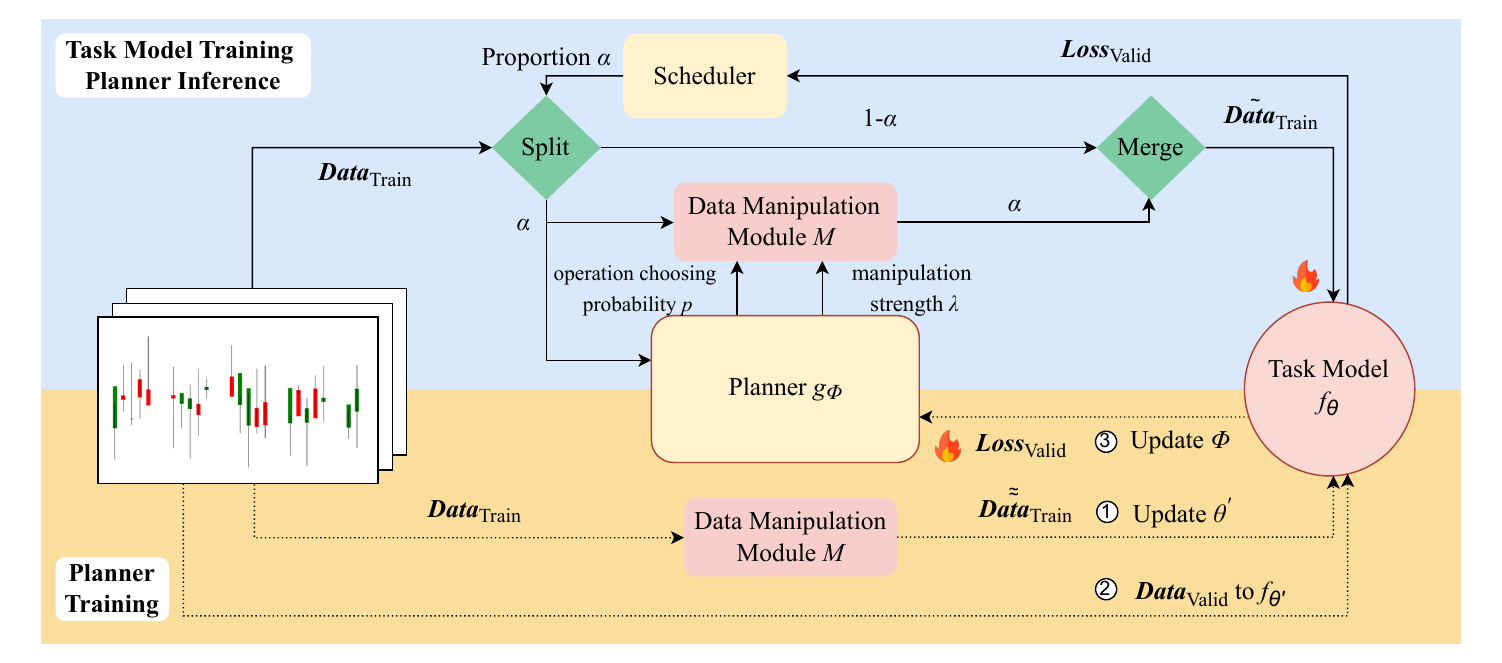}
    \caption{The workflow of training the planner and task model to learn a policy of controlling the data manipulation module with the validation loss of the task model. The training step of the planner is marked with (1), (2), (3), and $f_{\theta^{'}}$ is a copy of $f_{\theta}$. The fire icon marks the flow where parameters are updated.}
    \label{fig:overall_workflow}
\end{figure*}

\subsection{Observation of Concept Drift in Financial Data}
We investigate concept drift in financial data and compare it with time-series datasets: Weather\footnote{\url{https://www.bgc-jena.mpg.de/wetter/}}, Electricity \footnote{\url{https://archive.ics.uci.edu/dataset/321/electricityloaddiagrams20112014}}, and ETT \cite{zhou2021informer}. To examine the concept drift in $P_{t}(X,Y)=P_{t}(Y|X) \times P_{t}(X)$, we visualize $P_{t}(Y|X)$ and $P(X)$ respectively. For financial data, the $X$ comprises market prices and features, while the $Y$ is the next day's return of the close price. We use a t-SNE \cite{van2008visualizing} plot to mark training and testing data with different colors. Similarly, we use another t-SNE plot to visualize $P_{t}(X)$ over time. As shown in Fig.~\ref{fig:T-SNE_combined}, when compared to the stock price of McDonald's (MCD), the benchmark datasets exhibit a more overlapping distribution for both $P(Y|X)$ and $P(X)$. This indicates that the stock data demonstrate a more evident concept drift compared to other benchmarks.

\subsection{Validation–Test Proximity}
While the t-SNE visualization qualitatively illustrates concept drift between training and test sets, our method assumes that the validation data provide a closer approximation to the near-future test distribution than the historical training data. 
To verify this assumption, we perform a quantitative proximity analysis on both datasets using three statistical distance metrics—Population Stability Index (PSI), Kolmogorov–Smirnov (K–S) statistic, and Maximum Mean Discrepancy (MMD)—computed for Train–Test and Validation–Test pairs. The metrics are given by:

\begin{equation}
\mathrm{PSI} = \sum_{i=1}^{k} (p_i - q_i) \ln \left( \frac{p_i}{q_i} \right),
\end{equation}
where \( p_i \) and \( q_i \) denote the proportions of observations falling into bin \( i \) for the baseline and target distributions, respectively, and \( k \) is the total number of bins.

\begin{equation}
D_{\mathrm{KS}} = \sup_x \left| F_1(x) - F_2(x) \right|,
\end{equation}
where \( F_1(x) \) and \( F_2(x) \) are the empirical cumulative distribution functions (CDFs) of the two samples, and \( \sup_x \) denotes the maximum difference over all \( x \).

\begin{align}
\mathrm{MMD}^2(\mathcal{F}, U, V) 
&= \mathbb{E}_{u, u' \sim P_U}[k(u, u')] 
+ \mathbb{E}_{v, v' \sim P_V}[k(v, v')] \notag \\
&\quad - 2\,\mathbb{E}_{u \sim P_U, v \sim P_V}[k(u, v)],
\end{align}
where \( k(\cdot, \cdot) \) is a positive-definite kernel function, and \( P_U \) and \( P_V \) denote the two probability distributions being compared. In our experiments, we use the Radial Basis Function kernel.

\noindent
A higher PSI or MMD, or a larger K--S statistic, indicates a greater degree of distributional shift between the baseline and target samples.

We combine all market features with the target as a multi-variant time-series data. For the stocks dataset (daily, 2000–2024), we adopt a rolling-year protocol where each fold incrementally extends the training horizon by one year, splitting all samples from 2000 to $(2010 + k)$ into \textbf{Train/Validation/Test} subsets with ratios of 0.6/0.2/0.2. 
For the crypto dataset (hourly, 2023-09-27 to 2025-09-26), we apply a higher-frequency rolling protocol, expanding the training horizon by one month per fold while maintaining the same 0.6/0.2/0.2 chronological split. 
In each fold, PSI, K–S, and MMD are evaluated between the Train–Test and Validation–Test segments on feature distributions standardized using statistics computed from the training set.
Across both markets and temporal granularities, the results in Table \ref{tab:val_test_proximity} consistently show that $\text{Dist}(\text{Val}, \text{Test}) < \text{Dist}(\text{Train}, \text{Test})$, confirming that the validation window statistically resembles the near-future test distribution more closely than the historical training data. 
This quantitative evidence supports the use of validation feedback to guide our adaptive augmentation and planner updates in Section~\ref{Section:Adaptive Curriculum}.

\begin{table}[!t]
\centering
\caption{Average distributional distances between Train–Test and Validation–Test sets 
across two datasets. Lower values indicate that the validation distribution is 
closer to the test distribution than the distribution.}
\label{tab:val_test_proximity}
\vspace{-0.2cm}
\renewcommand{\arraystretch}{1.2}
\resizebox{1.0\linewidth}{!}{
\begin{threeparttable}
\begin{tabular}{l|ccc|ccc}
\toprule
\multirow{2}{*}{\textbf{Dataset}} &
\multicolumn{3}{c|}{\textbf{Train–Test}} &
\multicolumn{3}{c}{\textbf{Validation–Test}} \\
\cline{2-7}
 & \textit{PSI} & \textit{K–S} & \textit{MMD} & \textit{PSI} & \textit{K–S} & \textit{MMD} \\
\midrule
Stocks (2000–2024, daily) & 12.62 & 0.7415 & 0.7528 & \textbf{9.075} & \textbf{0.6367} & \textbf{0.6177} \\
Crypto (2023–2025, hourly) & 4.396 & 0.2988 & 0.1841 & \textbf{2.867} & \textbf{0.2680} & \textbf{0.1781} \\
\bottomrule
\end{tabular}
\end{threeparttable}
}
\end{table}

\subsection{Formulation of the Adaptive Control Objective} To mitigate the uncertainty caused by strong concept drift, we manipulate the training data to enhance generalization. Specifically, let $f_{\theta}(x)$ denote the task model. To best estimate the generalization ability of the model \cite{cheung2021adaaug}, the objective for learning the task model is:
\begin{equation}
\text{min} \quad \mathcal{L}_{\text{val}}(f_{\theta}, D_{\text{valid}}),
\label{eq:curriculum_objective}
\end{equation}

\noindent where $\mathcal{L}_{\text{val}}$ is the validation loss, and $D_{\text{valid}}$ is the validation dataset. Let $D_{\text{train}}$ denote the training dataset of the task model. Let $\mathcal{M}$ denote the manipulation module, we have $\tilde{x}_{\text{train}} \gets M(x_{\text{train}})$ for $x_{\text{train}} \in D_{\text{train}}$. 
Our objective is to develop an adaptive $\mathcal{M}$ that is simple yet effective in addressing the poor generalization caused by concept drift in financial data.
% Our objective is to develop an adaptive plug-and-play data manipulation workflow that is simple yet effective in addressing the poor generalization caused by concept drift in financial data.

\subsection{Augmentation Operations}
\label{section:augmentation}
We introduce the augmentation operations we will be using in our data manipulation module $\mathcal{M}$.

\noindent
\textbf{Single stock transformation operations.}
\begin{itemize}
\item \textbf{Robustness Enhancement:} Introduces controlled variations suited for noisy financial data. \textit{Jittering} adds noise to improve signal discrimination. \textit{Scaling} mitigates volatility-induced magnitude bias, and \textit{Magnitude Warping} \cite{um2017data} applies non-linear distortions common in price dynamics with cubic spline interpolation.
\item \textbf{Structural Variation:} \textit{Permutation} preserves local continuity but breaks strict sequencing, reflecting the partially stochastic nature of market evolution.
\item \textbf{Decomposition and Recombination:} \textit{STL Augmentation} \cite{cleveland1990stl} decomposes series into trend, seasonality, and residuals—components naturally present in financial time series—and bootstraps residuals to better model regime shifts and non-stationarity.
\end{itemize}

\noindent
\textbf{Multi-stock mix-up operations.}
\begin{itemize}
    \item \textbf{Segment Replacement Operations:} Replace segments of one stock with another, introducing local disruptions. \textit{Cut Mix} replaces a portion of one stock with another. 
    \item \textbf{Weighted Average Operations:} Combine data using weighted averages, creating smoother transitions between points. \textit{Linear Mix} linearly combines two stocks.
    \item \textbf{Frequency Domain Operations:} Combine underlying frequency components to capture cyclical patterns. \textit{Amplitude Mix} mixes the amplitude of the Fourier transform of two stocks which is essential for modeling cyclical patterns, such as seasonal effects and market cycles. Demirel and Holz (2024) combine significant frequencies of two stocks by mixing their phases and magnitudes.
\end{itemize}

\begin{figure*}[h!]
    \centering
    \includegraphics[width=\textwidth]{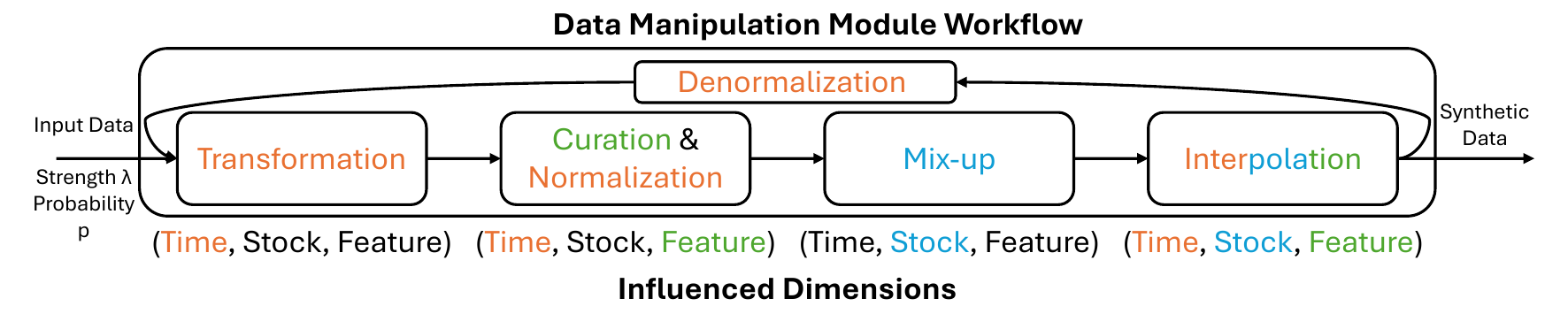}
    \caption{The proposed data manipulation module. The manipulated dimension of data is marked with the respective color.}
    \label{fig:data_manipulation_workflow}
\end{figure*}

\section{Method}
\label{method}

% {\color{purple}{
% Need to say something here. E.g., In this section, we propose a novel workflow to effectively address the problem presented in the previous section. First, we present the overall workflow. Then, we introduce the parameterized manipulation module. Finally, we present how to determine the optimal hyper-parameter values of the parameterized manipulation module by introducing a curriculum planner and over-fitting aware scheduler.

% In this section, we propose our novel workflow. First, we outline the overall workflow. Next, we introduce the parameterized manipulation module. Finally, we explain how the workflow determines the optimal policy to control the parameterized manipulation module by using a curriculum planner and an overfitting-aware scheduler. Our method operationalizes curation-aware synthesis: a parameterized manipulation module plus a planner–scheduler that adapts augmentation based on quality signals and logs full transformation provenance. 

In this section, we first describe the overall workflow, outlining how data is manipulated through the workflow. Next, we introduce the parameterized data manipulation module $\mathcal{M}$, which implements financially grounded operations to enhance data diversity while preserving realism and traceability. Finally, we detail the learning-guided controller, composed of a curriculum-based planner and an overfitting-aware scheduler, which continuously regulates manipulation strength and proportion of data to be manipulated based on validation feedback.
This controller closes the loop between data curation and model learning, enabling automated quality assurance and provenance tracking within a reproducible workflow.

\subsection{Overall Workflow}
To optimize the objective in equation (\ref{eq:curriculum_objective}) effectively with the guidance of gradients, we propose the workflow shown in Fig.~\ref{fig:overall_workflow}. The data manipulation module $\mathcal{M}$ generates augmented training samples $\tilde{x}_{\text{train}}$ with operation choosing probability matrix $p$ and manipulation strength parameter $\lambda$. To adaptively control $\mathcal{M}$, we introduce a trainable planner $g((f_{\theta}, x_{i}); \phi)$, which learns the policy $\pi_{\phi}(p, \lambda | f, x_{i})$ while the scheduler determines the proportion of data to be manipulated $\alpha$ using a heuristic algorithm. Additionally, we interleave task-model updates with planner updates on validation feedback, while provenance hooks (policy, probabilities, manipulation strengths, proportion of data to be manipulated) are persisted to enable exact replay.

In order to optimally control the data manipulation module $\mathcal{M}$ for each training sample $x_{\text{train}}$, inspired by AdaAug and MADAug, we formulate the learning of the adaptive augmentation curriculum as a bi-level optimization problem: 
\begin{equation}
 \begin{aligned}
	\min_{\phi} \quad & \mathcal{L}_{val}(f_{\theta}, x_{\text{valid}}), \: \: x_{\text{valid}}\in D_{\text{valid}}  \\
	 \textrm{s.t.} \quad & \theta= \arg\min_{\theta} \mathcal{L}_{train}(f_{\theta},\tilde{x}_{\text{train}}) \\
     \quad &  \tilde{x}_{\text{train}}=M(x_{\text{train}},\alpha,p,\lambda)
\end{aligned}
\label{eq:bi-level}
\end{equation}

\noindent where the planner and the task model are trained alternately with their respective objectives.

\subsection{Parameterized Data Manipulation Module}
\label{data_mm}

% {\color{red}Instead of just simpley collect the augmetation method and apply them to financial data , we designed a parameterized module as our adaptive augmetation method to the diverse financial data, this modules contain well designed methods to handle fiancnial data properly}

% The parameterized data manipulation module $\mathcal{M}$ is specifically designed to synthesize diverse and high-quality financial data. Given a training sample $x_{\text{train}}$ with the shape $(\text{Timestamps}, \text{Stocks}, \text{Features})$, the manipulation is guided by a probability matrix $p$ and a strength $\lambda$, as shown in Fig.~\ref{fig:data_manipulation_workflow}. For the $i$-th operation among $n$ single-stock operations and $j$-th operation among $m$ multi-stock mix-up operations, we have $\sum_{i=1}^{n}\sum_{j=1}^{m} p_{ij}=1$ and $\lambda_{ij} \in [0,1]$. The transformation and mix-up modules are designed to introduce diversity from different aspects, while curation, normalization, and interpolation modules ensure that the manipulated data remain reasonable and consistent with real financial data.

Unlike simply aggregating existing augmentation operations and applying them directly to financial data, the proposed data manipulation module $\mathcal{M}$ is designed as a \textbf{parameterized synthesis module} specifically tailored to the statistical and structural properties of financial time series. 
Each internal operation functions as a low-level augmentation primitive, while the \emph{method} lies in how these operations are integrated, parameterized, and coordinated by the operation choosing probability matrix $p$ and manipulation strength factor $\lambda$ to preserve financial validity while introducing realistic diversity. 
Rather than serving as a static augmentation toolkit, $\mathcal{M}$ provides a controllable mechanism for generating diverse yet high-fidelity financial data by accounting for temporal dependencies, cross-asset correlations, and market constraints such as K-line consistency and non-stationarity. 
Given a training sample $x_{\text{train}}$ with the shape $(\text{Timestamps}, \text{Stocks}, \text{Features})$, the manipulation is guided by $p$ and $\lambda$ as shown in Fig.~\ref{fig:data_manipulation_workflow}. 
For the $i$-th operation among $n$ single-stock transformations and the $j$-th operation among $m$ multi-stock mix-up operations, we have $\sum_{i=1}^{n}\sum_{j=1}^{m}p_{ij}=1$ and $\lambda_{ij}\in[0,1]$. 
The transformation and mix-up layers introduce controlled diversity from different market perspectives, while the curation, normalization, and interpolation layers enforce economic plausibility and maintain consistency with real-world financial dynamics. 
Financial time series exhibit strong temporal dependence, cross-asset co-movement, and price-based constraints (e.g., K-line consistency, non-stationarity, and heavy-tailed noise), and our module explicitly encodes these properties through four tightly coupled components, as illustrated in Fig.~\ref{fig:data_manipulation_workflow}. In our parameterized data manipulation module $\mathcal{M}$, each primitive is gated by financial integrity constraints and normalization/denormalization checks, turning augmentation into curated synthesis with auditable parameters.

\noindent\textbf{Transformation Layer.} We apply single-stock transformation operations as introduced in Section~\ref{section:augmentation} to the raw input feature to manipulate each stock feature independently. The operations are done with the data in its original value, as we will apply the following curation step in the original space to ensure fidelity. These operations are target-invariant to preserve the fidelity of financial data. Operations are parameterized with:

% \begin{itemize}
%     \item Robustness Enhancement: Introduce controlled variations to the data. \textit{Jittering} introduces randomness by adding noise, helping the model distinguish noise from signals. \textit{Scaling} adjusts the amplitude, reducing the model’s sensitivity to magnitude fluctuations. \textit{Magnitude Warping} applies non-linear distortions with cubic spline interpolation.
%     \item Structural and Temporal Variability: Modify the structural and temporal aspects of the data. \textit{Permutation} disrupts the global order of the data while maintaining local structures, helping the model focus on local patterns than the exact sequence of the price. 
%     \item Decomposition and Recombination: \textit{STL Augmentation} uses decomposition to separate the data into trend, seasonal, and residual components. Bootstrapping the residuals and recombining them handles non-stationarity.
% \end{itemize}
\begin{itemize}
    \item \textit{Jittering}: Manipulation strength $\lambda$ controls the standard deviation of the perturbation.
    \item \textit{Scaling}: $\lambda$ determines its amplitude.
    \item \textit{Magnitude Warping}~\cite{um2017data}: $\lambda$ controls its warp intensity.
    \item \textit{Permutation}: $\lambda$ controls how many parts the sequence is divided into before reordering.
    \item \textit{STL Augmentation}~\cite{cleveland1990stl}: $\lambda$ controls its period.
\end{itemize}

% \textit{Time Warping} alters the temporal aspect of the series, reflecting the variability in the speed at which financial events occur. This technique helps the model learn from sequences that may unfold over different time scales. 
% \textit{Inner Permutation} divides the time series into smaller segments and permutes each independently, aiding the model in adapting to short-term fluctuations.

% \begin{algorithm}[htbp]
% \caption{Mix-up Target Stock Sampling}
% \label{alg:mix-up_sampling}

% \begin{algorithmic}[1]

% \STATE\textbf{Input:} Source stock $a$, manipulation strength level $\lambda \in [0, 1]$, cointegration test p-values $\mathcal{P}$, number of top candidates $k$
% \STATE \textbf{Output:} Selected target stock index $b$

% \STATE Let the indices of the top \( k \) candidates be \( T \) and p-values of stocks in \( T \) be \( \mathcal{P}_T \).
% \STATE Apply a power transformation to \( \mathcal{P}_T \) to control skewness, yielding \( S \): \( S = \mathcal{P}_T^{\beta} \) for \( \lambda \leq 0.5 \) and \( S = \mathcal{P}_T^{1/\beta} \) for \( \lambda > 0.5 \), where \( \beta = 1 - \lambda \) if \( \lambda \leq 0.5 \) and \( \beta = \lambda \) if \( \lambda > 0.5 \).
% \STATE Normalize \( S \) to probabilities \( Q \) using softmax: \( Q = \frac{e^{S}}{\sum e^{S}} \), and sample \( b \) from \( T \) using \( Q \).

% \RETURN Selected target stock index $b$

% % \STATE \textcolor{purple}{As Bo said, Lines 3-5 should be explained in the text. Maybe in the paragraph \textbf{Mix-up Techniques}.}

% \end{algorithmic}
% \end{algorithm}

\begin{algorithm}[htbp]
\caption{Mix-up Target Stock Sampling}
\label{alg:mix-up_sampling}
\begin{algorithmic}[1]
\STATE \textbf{Input:} Source stock $a$; mix manipulation strength $\lambda\!\in\![0,1]$; cointegration $p$-values $\mathbf{p}\!=\!\{p_{aj}\}_j$; candidate count $k$
\STATE \textbf{Output:} Target stock index $b$
\STATE Exclude self: $p_{aa}\!\leftarrow\!\varnothing$; define candidate set $\mathcal{C}\!=\!\{j\!\ne\! a\mid p_{aj}\ \text{valid}\}$.
\STATE Set $\beta\!\leftarrow\!(1-\lambda)$ if $\lambda\!\le\!0.5$ else $\lambda$.
\STATE For each $j\!\in\!\mathcal{C}$, compute score
\[
S_j\!=
\begin{cases}
-\,p_{aj}^{\beta}, & \lambda\!\le\!0.5\ \text{(favor stronger cointegration)}\\[2pt]
p_{aj}^{1/\beta}, & \lambda\!>\!0.5\ \text{(favor weaker cointegration)}
\end{cases}
\]
\STATE Select top-$k$ indices $T$ by $S_j$ and apply a softmax over $\{S_j\}_{j\in T}$ to form probabilities $Q_j$.
\STATE Sample $b\!\sim\!\text{Categorical}(Q)$ and \textbf{return} $b$.
\end{algorithmic}
\end{algorithm}

\noindent\textbf{Curation and Normalization Layer.} After applying single-stock transformations, we curate the data to maintain financial consistency by setting the highest price feature to ``High" and the lowest to ``Low". To conduct mix-up operations which involve the exchange of data between stocks, it is crucial to normalize the data. This is achieved using rolling-window standard normalization applied to each feature of each stock. For multiple layers of manipulation, the data is denormalized when reintroduced to the workflow.

% algo v1
% We select a target stock $b$ by identifying the top $k$ most statistically correlated stocks with the lowest cointegration test p-values. By applying a transformation based on the manipulation strength parameter $\lambda$, we control the skewness of the sampling distribution, where a larger manipulation strength increases the probability of selecting stocks that are less tightly cointegrated with the source stock, bringing strong modification.

\noindent\textbf{Mix-up Layer.} We apply multi-stock mix-up operations as introduced in Section~\ref{section:augmentation} to the normalized data to mix each source stock $a$ with target stock $b$, as described in Algorithm \ref{alg:mix-up_sampling}. We select a target stock $b$ by identifying the top $k$ most correlated stocks with the source stock $a$ based on cointegration test p-values. To control the skewness of the operation choosing probability, we apply a transformation to these p-values, adjusted by a manipulation strength parameter $\lambda$. For $\lambda\leq0.5$, a power transformation compresses the p-values, favoring more cointegrated stocks. For $\lambda>0.5$, an inverse power transformation expands the p-values, increasing the chance of selecting less cointegrated stocks. The transformed $S$ are normalized to $Q$, and the target stock $b$ is sampled from this distribution. The mix-up methods are target-variant because they involve creating new samples by combining both the features and targets of two different stocks. Operations are parameterized with:
\begin{itemize}
    \item \textit{Cut Mix}: $\lambda$ determines the area of the patch replaced between two samples
    \item \textit{Linear Mix}: $\lambda$ controls the interpolation ratio between the two inputs and their labels
    \item \textit{Amplitude Mix}: $\lambda$ it adjusts the relative amplitude or energy contribution of each signal.
    \item \textit{Demirel and Holz (2024)}: $\lambda$ determines the blending ratio between the two signals.
\end{itemize}

% \textit{Binary Mix} randomly selects segments from either of the two stocks based on a binomial distribution.
%\begin{algorithm}[htbp]
%\caption{Interpolation Compensate}
%\label{alg:interpolation_compensate}
%\begin{algorithmic}[1]

%\STATE\textbf{Input:} Original Data \( \mathbf{x} \), augmented data \( \mathbf{y} \), Factor \( b_{\text{max}} \)
%\STATE\textbf{Output:} Compensated data \( \mathbf{x'} \) 

%\STATE Randomly select feature \( k \) for fast estimation

%\STATE Calculate mutual information \( \text{MI}_{xy} \) and baseline \( %\text{MI}_{xx} \) to measure similarity and maximum similarity

%\STATE Compute factor 
% $ b_{\text{mix}} = b_{\text{max}} - \left( \frac{\text{MI}_{xy}}{\text{MI}_{xx}} \right) b_{\text{max}}$

% \STATE Compute compensated data $ \mathbf{x^{'}} = b_{mix} \mathbf{x} + (1 - b_{mix}) \mathbf{y} $

%\RETURN compensated data $ \mathbf{x'} = b_{\text{mix}} \mathbf{x} + (1 - b_{\text{mix}}) \mathbf{y} $

%\end{algorithmic}
%\end{algorithm}

\noindent\textbf{Interpolation Compensation Layer.} While the curation module sets hard constraints to maintain the fidelity of the financial data, we also apply an interpolation compensation to mitigate potential extreme samples. We propose a mutual-information-aware mixing strategy termed \textit{Binary Mix}. Unlike standard interpolation methods that blend two samples uniformly or randomly, Binary Mix adaptively adjusts the interpolation ratio based on the similarity between the original and augmented data. Specifically, we compute the mutual information defined in Equation \ref{Mutualinfo}, between the two samples to estimate how semantically aligned they are, and reduce the mixing weight accordingly. This ensures that less similar augmentations contribute less to the final sample, preserving task-relevant structure. The procedure is detailed in Algorithm~\ref{alg:binarymix} where the factor $b_{\text{mix}}$ is calculated with the mutual information. The less mutual information the augmented data has with the original data, the more it is compensated with the original data.
% \begin{equation}
% \mathrm{MI}(X; Y) = \sum_{x \in \mathcal{X}} \sum_{y \in \mathcal{Y}} 
% p(x, y) \log \left( \frac{p(x, y)}{p(x) \, p(y)} \right),
% \label{Mutualinfo}
% \end{equation}

\begin{equation}
\mathrm{MI}(X; Y) = 
\iint_{\mathcal{X} \times \mathcal{Y}} 
f_{X,Y}(x, y) \,
\log \left( 
\frac{f_{X,Y}(x, y)}{f_X(x) \, f_Y(y)} 
\right)
\, \mathrm{d}x \, \mathrm{d}y,
\label{Mutualinfo}
\end{equation}

\noindent where \( X \) and \( Y \) are random variables with joint distribution \( p(x, y) \) and marginals \( p(x) \) and \( p(y) \).

\begin{algorithm}[htbp]

\caption{Binary Mix: randomly selects segments from either of the two stocks based on a binomial distribution.}
\label{alg:binarymix}
\begin{algorithmic}[1]

\STATE\textbf{Input:} Original Data \( \mathbf{x} \), augmented data \( \mathbf{y} \), Factor \( b_{\text{max}} \)
\STATE\textbf{Output:} Compensated data \( \mathbf{x'} \) 

\STATE Randomly select feature \( k \) for fast estimation

\STATE Calculate mutual information \( \text{MI}_{xy} \) and baseline \( \text{MI}_{xx} \) to measure similarity and maximum similarity

\STATE Compute factor 
 $ b_{\text{mix}} = b_{\text{max}} - \left( \frac{\text{MI}_{xy}}{\text{MI}_{xx}} \right) b_{\text{max}}$

\STATE Compute compensated data $ \mathbf{x^{'}} = b_{mix} \mathbf{x} + (1 - b_{mix}) \mathbf{y} $

\RETURN compensated data $ \mathbf{x'} = b_{\text{mix}} \mathbf{x} + (1 - b_{\text{mix}}) \mathbf{y} $

\end{algorithmic}
\end{algorithm}

\begin{algorithm}[htbp]
\caption{Proportion $\alpha\ $ Scheduler}
\label{alg:split_rate_adjustment}
\begin{algorithmic}[1]
\STATE \textbf{Input:} Current early stopping counter $C_{\text{es}}$, Last early stopping counter $C_{\text{les}}$, Epoch $E$, Threshold $\tau$
\STATE \textbf{Output:} Proportion of data to be manipulated $\alpha$
\STATE Let rate penalty $R_{\text{penalty}} = 1 \text{ if } C_{\text{es}} > C_{\text{les}} \text{ else } 0.1$.
\STATE Update $C_{\text{les}} = C_{\text{es}}$.
\RETURN $\alpha = \min(\tanh(\frac{E}{\tau}) + 0.01, 1.0) \times R_{\text{penalty}}$
\end{algorithmic}
\end{algorithm}

\subsection{Adaptive Curriculum}
\label{Section:Adaptive Curriculum}

% Given the diversity of financial data, tasks, and models, finding a generalized policy to control the parameterized data manipulation model $\mathcal{M}$ is challenging. We introduce an adaptive planner that observes both model and data states to optimally control $\mathcal{M}$, while an overfitting-aware scheduler integrates augmented data into training via a dynamic curriculum.

Given the heterogeneity of financial data, learning objectives, and model architectures, 
designing a unified policy to control the parameterized data manipulation module $\mathcal{M}$ remains non-trivial. 
We introduce an \emph{adaptive planner} that jointly observes model and data states to determine optimal parameters, 
while an \emph{overfitting-aware scheduler} progressively integrates augmented samples through a dynamic curriculum.

% We introduce the adaptive planner, which learns the optimal policy to control $\mathcal{M}$ by observing the state of both the model and the data. At the same time, the scheduler integrates the augmented data into task model training through an overfitting-aware curriculum.

% maybe too detail we want high level story here so not stating the fact but narrating the reasons
% \subsection{Problem Formulation}

% We aim to provide optimal control of the data manipulation module for each training sample $x_{\text{train}}$. Following AdaAug \cite{cheung2021adaaug} and MADAug \cite{hou2023learn}, we formulate the objective of the curriculum as:
% \noindent\begin{equation}
% \underset{\alpha,p,\lambda}{\text{min}} \quad \mathcal{L}_{val}(f_{\theta},D_{\text{valid}})
% \label{eq:crriculum_objective}
% \end{equation}

% We aim to minimize the validation loss $\mathcal{L}_{val}$ given the task model $f(x;\theta)$, and the validation data $D_{\text{valid}}$, while controlling the proportion of the data to be manipulated $\alpha$, the operation choosing probability of choosing operations $p$ and manipulation strength $\lambda$. For the i-th operation among $n$ single-stock operations and j-th operation among $m$ multi-stock mix-up operations, we have $\sum_{i=1}^{n}\sum_{j=1}^{m} p_{ij}=1$ and $\lambda_{ij} \in [0,1]$.

\noindent\textbf{Curriculum Planner.} We train a planner $g_{\phi}$ to learn the policy $\pi_{\phi}(p,\lambda|f,x_{i})$ to optimize the objective function in equation (\ref{eq:bi-level}). The state includes both the task model $f_{\theta}$ and the input sample $x_{i} \in D$, ensuring that the curriculum can be determined based on both the model and the data, as supported by findings in \cite{saxena2019data}. To efficiently represent the state, we use high-level representations of $f_{\theta}$ and $x_{i}$. For the state of the model, we extract features from the second-to-last fully connected layer inserted into the task model. Comparably, for the state of the input sample $x_{i}$, we compute key metrics such as mean, volatility, momentum, skewness, kurtosis, and trend. More concretely, we calculate: $\text{Momentum} = X_{\text{last}} - X_{\text{first}},
$
where \( X_{\text{last}} \) and \( X_{\text{first}} \) are the values of the feature at the last and first observations in the window, respectively.
%\begin{equation}
%\%text{Moving Average (MA)} = \frac{1}{n} \sum_{i=1}^{n} X_i,
%\end{equation}
%where \( n \) is the length of the rolling window and \( X_i \) is the value of the feature at observation \( i \).
%\begin{equation}
%\text{Volatility } (\sigma) = \sqrt{\frac{1}{n-1} \sum_{i=1}^{n} (X_i - \overline{X})^2},
%\end{equation}
%where \( X_i \) is the value of the feature at observation \( i \), \( \overline{X} \) is the mean of the feature values in the window, and \( n \) is the number of observations in the window.
$
\text{Skewness} = \frac{\frac{1}{n} \sum_{i=1}^{n} (X_i - \overline{X})^3}
{\left( \frac{1}{n} \sum_{i=1}^{n} (X_i - \overline{X})^2 \right)^{3/2}},
$
where \( X_i \) is the value of the feature at observation \( i \), \( \overline{X} \) is the mean of the feature values, and \( n \) is the number of observations in the window.
$
\text{Kurtosis} = \frac{\frac{1}{n} \sum_{i=1}^{n} (X_i - \overline{X})^4}
{\left( \frac{1}{n} \sum_{i=1}^{n} (X_i - \overline{X})^2 \right)^{2}} - 3,
$
where \( X_i \) is the value of the feature at observation \( i \), \( \overline{X} \) is the mean of the feature values, and \( n \) is the number of observations in the window.
$
\text{Trend} = \frac{\sum_{i=1}^{n} (t_i - \overline{t})(X_i - \overline{X})}
{\sum_{i=1}^{n} (t_i - \overline{t})^2},
$
where \( t_i \) is the time or index of observation \( i \), \( X_i \) is the value of the feature at observation \( i \), and \( \overline{t} \) and \( \overline{X} \) are the means of the time and feature values within the window, respectively.

This ensures that the planner $g_{\phi}$ can effectively learn a policy that adapts to the model and data, thereby optimizing the curriculum more efficiently.

% During training, the planner $g_{\phi}$ and task model $f_{\theta}$ are optimized jointly with the bi-level optimization objective:

% \begin{equation}
%  \begin{aligned}
% 	\min_{\phi} \quad & \mathcal{L}_{val}(f_{\theta}, x_{\text{valid}})  \\
% 	 \textrm{s.t.} \quad & \theta= \arg\min_{\theta} \mathcal{L}_{train}(f_{\theta},\tilde{x}_{\text{train}}) \\
%      \quad &  \tilde{x}_{\text{train}}=M(x_{\text{train}},\alpha,p,\lambda)
% \end{aligned}
% \label{eq:bi-level}
% \end{equation}
% where the manipulated training data sample  $\tilde{x}_{\text{train}}$ is synthesized by the workflow as displayed in Fig.~\ref{fig:overall_workflow}.

To address risk from inference uncertainty in financial tasks, we propose a loss function inspired by the Sharpe ratio, incorporating standard deviation to penalize volatility:
\begin{equation}
\mathcal{L}=E(loss)+\gamma \times \sigma(loss)
\end{equation}
This formulation guides the model away from risky inferences by penalizing. We set $\gamma=0.05$ for experiments.

% \noindent\textbf{Over-fitting Aware Scheduler.} While the planner controls the operation probability $p$ and manipulation strength $\lambda$, the proportion of data to be manipulated $\alpha$ is determined by the scheduler to provide a reasonable curriculum. In computer vision research, it is common to apply augmentation to all training samples \cite{cubuk2018autoaugment, cubuk2020randaugment}. However, such a fixed application of augmentation may not be suitable for our objective of addressing uncertainty in financial markets. Therefore, we propose a dynamic data manipulation strategy, as outlined in Algorithm \ref{alg:training_scheme}. The principles of curriculum learning is to let the model progressively learns from easier to harder samples. As synthic data introduce addtional diversity to the original dataset, the larger to porportion of the synthic data, the samples would be harder, so the $\alpha$ is increased with the epoch. However, excessive manipulation of the data may adversely affect the learning process. As our objective is to mitigate concept drift, which can be indicated by the overfitting occurs when the model performs well on training data but poorly on unseen validation set, we avoid the over-manipulation by monitering the overfitting indicators in the training process. Overfitting can be monitored by tracking the validation loss: if the validation loss in the current epoch is not lower than in the previous epoch by a specified threshold.

\noindent\textbf{Over-fitting Aware Scheduler.} While the planner controls the operation choosing probability $p$ and manipulation strength $\lambda$, the proportion of data to be manipulated $\alpha$ is determined by the scheduler to provide a reasonable curriculum. In CV research, it is common to apply augmentation to all training samples \cite{cubuk2018autoaugment, cubuk2020randaugment}. However, such a fixed application of augmentation may not be suitable for our objective of addressing uncertainty in financial markets. Therefore, we propose a dynamic data manipulation strategy, as outlined in Algorithm \ref{alg:split_rate_adjustment}. Curriculum learning encourages the model to progress from simpler to more challenging examples. As data augmentation increases sample diversity and complexity, the poportion of augmented data can be gradually raised during training. Hence, $\alpha$ is designed to increase over epochs, reflecting a soft curriculum rather than a fixed rule. However, excessive manipulation may adversely affect the learning process. Our objective is to mitigate poor generalization caused by concept drift, which is often indicated by overfitting—where the model performs well on the training data but poorly on unseen validation data. To prevent over-manipulation of the data, we monitor overfitting during training by the validation loss: if the validation loss in the current epoch is not lower than in the previous epoch by a specified threshold, it suggests potential overfitting. When this occurs, the penalty to regulate frequent manipulation $R_{\text{penalty}}$ is removed to introduce more diverse data.

% By scheduling the proportion of manipulated data, we keep a balance between enhancing generalization and maintaining the training robustness, improving the model's ability to handle uncertainty in financial markets without introducing unnecessary complexity.
By scheduling the proportion of data to be manipulated, we balance generalization and robustness, enhancing the model’s ability to handle market uncertainty without added complexity.

\begin{algorithm}[htbp]
\caption{Joint Training Scheme}
\label{alg:training_scheme}

\begin{algorithmic}[1]

\STATE \textbf{Input:} Training set $D_{\text{train}}$; 
Validation set $D_{\text{valid}}$;
Update frequency $freq$

\STATE \textbf{Output:} Planner $g_{\phi}$;  task model $f_{\theta}$

\STATE Initialize $\theta_0$, $\phi_0$

\WHILE{max epoch not reached and no early stopping}
    % \STATE Sample batch $x_{\text{train}}$ from $\mathcal{D}_{train}$
    \STATE Get $\alpha$ with Algo \ref{alg:split_rate_adjustment} and $p,\lambda=g(f_{\theta},x_{\text{train}})$
    \STATE Update $f_{\theta}(\tilde{x}_{\text{train}}=\mathcal{M}(\alpha,p,\lambda,x_{\text{train}});L_{\text{train}})$

    % add an inner loop of every $freq$ steps
    \IF{current step is divisible by $freq$}
        % \STATE Sample mini-batch $x_{\text{valid}}$ from $\mathcal{D}_{valid}$
        \STATE Copy $f_{\theta}$ as $f_{\theta^{'}}$, update it with $\tilde{\tilde{x}}_{\text{train}}$ with eq.(\ref{equation:x_valid})
        \STATE Update $g_{\phi}(x_{\text{valid}};L_{\text{valid}}(f_{\theta^{'}}))$ 
    \ENDIF
\ENDWHILE
\STATE \textbf{return} Trained planner $g_{\phi}$; trained task model $f_{\theta}$
\end{algorithmic}
\end{algorithm}

\noindent\textbf{Planner Training Scheme.} Given the complexity of this bi-level optimization problem, we adopt methods from \cite{shu2019meta, hou2023learn} to jointly train the planner $g_{\phi}$ and the task model $f_{\theta}$. The task model is trained within the training loop, while the planner is trained within the validation loop. This relies on a local temporal smoothness assumption: when the data are partitioned chronologically, the validation segment is statistically closer to the upcoming test segment than the earlier training window as supported by Section \ref{section:problem_formulation}. This proximity supports using validation risk as a practical surrogate for test risk, thereby grounding the bi-level planner–scheduler optimization in a measurable temporal relationship. The alternating update scheme is outlined in Algorithm \ref{alg:training_scheme}. The task model $f_{\theta}$ is first updated. To stabilize the training process, the planner will then be trained for every frequency $freq$ steps to learn the policy $\pi_{\phi}(p,\lambda|f_{\theta},x_{i})$, with the validation loss of $f_{\theta^{'}}$, which is a copy of $f_{\theta}$ at the current step.

% The explore/exploit scheme
%removed this and combined upwards
%To stabilize the training process, we control the update of the planner by a frequency $freq$ where the planner $g_{\phi}$ is updated once for every $freq$ updates of the task model $f_{\theta}$. While the task model $f_{\theta}$ is trained with its original objective, the planner learns the policy $\pi_{\phi}(p,\lambda|f,x_{i})$ with the validation loss of $f_{\theta}$ on $x_{\text{valid}}$.
To learn the operation choosing probability of operation $p$, we
generate all the augmented data of the $n \times m$ sets of the operations with their manipulation strength $\lambda_{ij}$ and sum all the augmented data up with their weight in the probabilistic $p_{ij}$, where 
\begin{equation}
\tilde{\tilde{x}}_{\text{train}}= \sum^{n}_{i=1}\sum^{m}_{j=1}p_{ij}\times \mathcal{M}(1,1,\lambda_{ij},x_{\text{train}}) .
\label{equation:x_valid}
\end{equation}
Using a weighted sum instead of sampling with $p_{ij}$ accounts for the effect of every augmentation operation combination, thus leading to a better estimation when updating $\phi$.

To deal with non-differentiable operations, we use a straight-through gradient estimator \cite{li2020differentiable} to optimize the manipulation strength $\lambda$, where  $\frac{\partial M(x_{i}) }{\partial \lambda}=1$, with gradient:
\begin{equation}
\begin{aligned}
\frac{\partial \mathcal{L^{\text{val}}}}{\partial 
\lambda}&= \sum_{i}p_{ij}
     \frac{\partial \mathcal{L^{\text{val}}}}{\partial f_{\theta}}
     \frac{\partial f_{\theta}}{\partial M(x_{i})}
     \frac{\partial M(x_{i}) }{\partial \lambda}
\\&= \sum_{i}p_{ij}
     \frac{\partial \mathcal{L^{\text{val}}}}{\partial f_{\theta}}
     \frac{\partial f_{\theta}}{\partial M(x_{i})}.
\end{aligned}
\label{eq;lambda}
\end{equation}
The outer loop updates of planner $\phi_t$ is given by:
\begin{equation}
\phi_{t+1} = \phi_t - \beta \frac{1}{n^{\text{val}}} 
\sum_{i=1}^{n^{\text{val}}} 
\nabla_{\phi} \mathcal{L}^{\text{val}}_i(\hat{\theta}_t(\phi_t))
\label{eq:outer_update}
\end{equation}

where $\beta$ is the learning rate and $\hat{\theta}$ is the task model.

% \begin{figure}[htbp]
%     \centering
%     \includegraphics[width=0.8\linewidth]{figs/training_loss.pdf}
%     \caption{The training and validation loss curve w/wo our workflow applied.}
%     \label{fig:training_loss}
% \end{figure}

% \begin{figure}[h!]
%     \centering
%     \begin{subfigure}{1\columnwidth}
%         \centering
%         \parbox[c][3cm][c]{1\linewidth}{ % Adjust the 6cm to the same height as above
%             \includegraphics[width=\linewidth]{figs/dqn_intc_original.png}
%         }
%         % \captionsetup{font=scriptsize}
%         \caption{DQN on INTC}
%         \label{fig:MDM_adjclose}
%     \end{subfigure}%
%     \vspace{0.1cm}
%     \hfill
%     % \label{fig:MDM_result}
%         \begin{subfigure}{1\columnwidth}
%         \centering
%         \parbox[c][3cm][c]{1\linewidth}{ % Adjust the 6cm to an appropriate height
%             \includegraphics[width=\linewidth]{figs/dqn_intc_madaug.png}
%         }
%         % \captionsetup{font=scriptsize}
%         \caption{DQN + Ours on INTC}
%         \label{fig:MDM_TSNE}
%     \end{subfigure}%
%     \caption{Trading results where buy and sell actions are marked with green and red labels.}
%     \label{fig:Trading_result}
% \end{figure}

% \input{contents/experiment}
\begin{table*}[htbp]
\caption{Unified comparison over \textbf{Stocks} (upper block) and \textbf{Cryptos} (lower block) across five model families.}
\centering
\setlength{\tabcolsep}{4.5pt} % a bit tighter, still readable
\footnotesize
\begin{threeparttable}
\begin{tabular}{l|
ccc|ccc|ccc|ccc|ccc}
\toprule
& \multicolumn{3}{c|}{\textbf{GRU}} & \multicolumn{3}{c|}{\textbf{LSTM}} & \multicolumn{3}{c|}{\textbf{DLinear}} & \multicolumn{3}{c|}{\textbf{TCN}} & \multicolumn{3}{c}{\textbf{Transformer}} \\
\textbf{Method} & \textit{MSE} & \textit{MAE} & \textit{STD} & \textit{MSE} & \textit{MAE} & \textit{STD} & \textit{MSE} & \textit{MAE} & \textit{STD} & \textit{MSE} & \textit{MAE} & \textit{STD} & \textit{MSE} & \textit{MAE} & \textit{STD} \\
\midrule
\multicolumn{16}{l}{\textbf{Stocks} \;\; (exponents: MSE $\times 10^{-4}$, MAE $\times 10^{-2}$, STD $\times 10^{-3}$)} \\
\midrule
Original        & 22.76 & 3.388 & 5.140 & 5.070 & 1.578 & 1.664 & 52.15 & 5.501 & 9.601 & 40.44 & 4.614 & 10.01 & 8.608 & 2.216 & 1.915 \\
RandAug         & 16.74 & 2.864 & 3.994 & 4.646 & 1.495 & 1.613 & 662.7 & 19.86 & 116.7 & 58.74 & 4.925 & 22.12 & 8.264 & 2.160 & 1.892 \\
TrivialAug      & 15.62 & 2.670 & 4.114 & 4.827 & 1.536 & 1.634 & 571.6 & 18.40 & 100.7 & 46.83 & 5.142 & 10.79 & 7.474 & 2.041 & 1.814 \\
AdaAug          & 17.64 & 2.972 & 3.995 & 4.791 & 1.538 & \textbf{1.567} & 22.62 & 3.803 & 3.234  & 49.61 & 5.203 & 11.61 & 7.602 & 2.062 & 1.819 \\
Ours            & \textbf{13.14} & \textbf{2.496} & \textbf{3.366} & \textbf{4.253} & \textbf{1.410} & 1.571 & \textbf{4.550} & \textbf{1.485} & \textbf{1.559} & \textbf{34.87} & \textbf{4.431} & \textbf{7.864} & \textbf{7.471} & \textbf{2.046} & \textbf{1.795} \\
\midrule
\multicolumn{16}{l}{\textbf{Cryptos} \;\; (exponents: MSE $\times 10^{-5}$, MAE $\times 10^{-3}$, STD $\times 10^{-3}$)} \\
\midrule
Original        & 8.144 & 6.551 & 8.975 & 4.426 & 4.568 & 6.453 & 4.291 & 4.428 & 6.339 & 913.3 & 29.57 & 90.95 & 8.356 & 6.726 & 7.555 \\
RandAug         & 8.315 & 6.442 & 9.042 & 4.348 & 4.509 & 6.385 & 4.275 & 4.412 & 6.324 & 1883  & 40.94 & 125.5 & 7.295 & 6.400 & 7.232 \\
TrivialAug      & 8.268 & 6.571 & 8.967 & 4.361 & 4.518 & 6.401 & 4.273 & 4.411 & 6.322 & 855.2 & 32.93 & 83.30 & 11.41 & 7.318 & 8.809 \\
AdaAug          & 12.54 & 8.142 & 11.16 & 4.460 & 4.583 & 6.476 & 4.278 & 4.422 & 6.329 & 332.3 & 32.43 & 56.38 & 7.877 & 6.484 & 7.395 \\
Ours w/o mixup  & 7.552 & 6.231 & 8.437 & 4.402 & 4.511 & 6.422 & 4.252 & 4.403 & 6.278 & 469.3 & 29.47 & 66.37 & 7.732 & 6.207 & 7.113 \\
Ours            & \textbf{6.916} & \textbf{5.816} & \textbf{8.157} & \textbf{4.235} & \textbf{4.324} & \textbf{6.210} & \textbf{4.138} & \textbf{4.374} & \textbf{6.122} & \textbf{307.7} & \textbf{31.87} & \textbf{53.73} & \textbf{4.941} & \textbf{4.982} & \textbf{6.661} \\
\bottomrule
\end{tabular}

\end{threeparttable}

\label{tab:main_results}
\end{table*}

\section{Experiments}
%FIRST VERSION
%The purpose of our experiment is to evaluate the adaptability and robustness of our proposed %workflow across various financial tasks and models. To achieve this, we structured our %experiments around the following research questions:

% We want to testify if our method address the challenges of (1)lack of appropriate augmentation methods for financial data (2) Lack of an adaptive method for adapting such augmentations across various models and tasks. (3) the concept drift that worsen out-of-sample model performance. As such, we conduct experiments to first determine the effectiveness of the augmentations that are tailored for financial data. Subsequently, we evaluate the effectiveness of our adaptive and task agnostic framework. The main questions we want to answer are:
% \begin{enumerate}
%     \item Are the augmentation operations we proposed suitable?
%     \item Can our workflow adapt to different models and tasks?
%     \item Does our workflow address concept drift?
% \end{enumerate}
% The experiments were conducted in a way to evaluate the effectiveness and capability of each step of the proposed workflow, from augmentations to the data manipulation module, the planner and finally task transfer. 
%thumbsup
% yeah I just rephrase it

We aim to evaluate whether our adaptive dataflow system effectively addresses three key challenges:
\begin{enumerate}
\item \textbf{Workflow effectiveness.}
Does the proposed dataflow system improve downstream task robustness and overall data-processing efficiency compared to existing augmentation and generation pipelines?

\item \textbf{System adaptability.}  
Can the workflow generalize across heterogeneous models and tasks?

\item \textbf{Data fidelity and curation quality.}  
Does the pipeline generate diverse yet realistic financial time-series data for downstream analysis?

\end{enumerate}

% We aim to evaluate whether our method addresses three key challenges: i) the lack of effective financial data augmentations; ii) the absence of adaptive methods across models and tasks; and iii) concept drift that degrades performance. Main research questions are:
% \begin{enumerate}
%     % \item Are the augmentation operations we proposed suitable for financial data?
%     \item Does our workflow lead to better downstream performance?
%     \item Can our workflow adapt to different models and tasks?
%     \item Do we get high-quality financial data from our pipeline?
% \end{enumerate}

% The experiments were designed to evaluate the effectiveness and capabilities of each step in the proposed workflow, from augmentations to the data manipulation module, the planner, and finally task transfer.

\subsection{Experiment Setup}
%Dataset part
\paragraph{Datasets} 
We evaluate on two real-world financial markets:
(i) \textbf{Stocks (daily).} Price and technical indicators for 27 stocks of the Dow Jones Industrial Average from \textbf{2000-01-01} to \textbf{2024-01-01};
(ii) \textbf{Crypto (hourly).} Price and technical indicators for BTC, ETH, DOT, and LTC from \textbf{2023-09-27} to \textbf{2025-09-26}. The datasets were divided into training, validation, and test sets with ratios of $0.6$, $0.2$, and $0.2$. All normalization statistics, cointegration measures, and other rolling or windowed computations were strictly estimated within the training set to prevent any temporal leakage. Experiments were run on a GeForce RTX 4090 GPU.\\
%Benchmark part
\paragraph{Benchmark for Augmented Data}
We compared our augmented data with data generated by representative time-series generative models, including TimeGAN, SigCWGAN \cite{ni2020conditional}, RCWGAN \cite{he2022novel}, GMMN \cite{li2015generative}, CWGAN \cite{yu2019cwgan}, RCGAN \cite{esteban2017real} and Diffusion-TS. \\
\paragraph{Benchmark for The Whole Workflow} In the experiments, the proposed method is compared with the following methods: (1) \textbf{Original}: Uses the original data without any manipulation to the training scheme. (2) \textbf{RandAugment}: Employs randomly chosen augmentations where $\lambda=1$ for each augmentation and the proportion $\alpha=0.5$ throughout the training process. (3) \textbf{TrivialAugment}: This baseline employs randomly chosen augmentations with a randomized $\lambda$ and fixed proportion $\alpha=0.5$. (4) \textbf{AdaAug}: This setup utilizes the planner $g_{\phi}$ for augmentations but maintains an $\alpha=0.5$. 
\paragraph{Task Model} 
We evaluate our method using five representative architectures across different model families: GRU, LSTM, DLinear, TCN, and Transformer on forecasting tasks, and DQN and PPO for reinforcement learning–based trading tasks.
To integrate any model into our method, the downstream task model simply has to adopt a two-stage architectural pattern that explicitly separates feature extraction from prediction tasks. Unlike traditional models that employ a mapping from input sequences to predictions, we implement a modular design consisting of two distinct functions: the feature extraction function \( j(\cdot) \) that extracts learned representations from the second-to-last layer of the network, and the prediction function \( k(\cdot) \) that maps these extracted features to final predictions where \( k(j(\cdot)) \) is the typical forward function. For models that do not have at least 2 fully-connected layers, we add them to facilitate this process. For the planner, we use a Transformer. For all models, we set the learning rate to be 0.001, and the planner input dimensions to be 128. The batch size for GRU and TCN at set at 128, LSTM and Transformer at 256, and DLinear at 1024. The patience for GRU, LSTM, TCN and Transformer is set at 5 while DLinear at 8. The hidden dimensions for the task model is set at 512 for GRU, LSTM, DLinear and TCN, and at 256 for Transformer. The planner layers are set at 2 for GRU, LSTM, TCN and Transformer, and at 5 for DLinear, while the planner hidden dimensions are set at 256 for GRU, LSTM, DLinear and TCN, and at 128 for Transformer.
The key hyperparameters are as follows:
\begin{itemize}
    \item \textbf{Threshold $\tau$}: This has the same concept as the temperature parameter, and affects the proportion of data to be manipulated $\alpha$. We set $\tau_{GRU}$ to 14,  $\tau_{LSTM}$, $\tau_{TCN}$ and $\tau_{Transformer}$ to 5, and $\tau_{DLinear}$ to 30.
    \item \textbf{Frequency $freq$}: This controls the frequency in which the planner updates. A smaller $freq$ value would increase the number of updates, which increases the time taken for training but helps the planner learn better (and vice versa). We set $freq_{GRU}$ to 15, $freq_{LSTM}$ and $freq_{Transformer}$ to 5, $freq_{DLinear}$ to 8, and $freq_{TCN}$ to 10.
    \item \textbf{Planner training epoch start}: This controls when the planner starts learning. A lower start would allow the planner to update earlier, which improves the quality of the synthetic time series but increase the time taken for training. We set the start epoch to 10 for GRU, 5 for LSTM, and 2 for DLinear, TCN and Transformer.
\end{itemize}

\paragraph{Reinforcement Learning Environment}
To evaluate the transferability of the adaptive planner in downstream trading, 
we implement a single-asset discrete-action trading environment. 
The environment simulates realistic trading dynamics with transaction costs and evolving net value.

\textbf{Action space, position, and valuation.}
We use an all-in/all-out regime with discrete actions
$a_t\in\{-1,0,1\}$ for \emph{sell}, \emph{hold}, \emph{buy}.
Let $s_t\ge 0$ denote the number of shares held at $t$,
$\text{cash}_t$ the cash balance, $P_t$ the adjusted close,
and $V_t=\text{cash}_t+s_t P_t$ the portfolio value.
A proportional transaction cost $c=10^{-3}$ applies to traded notional.

\emph{Execution:}
If $a_t=1$ (enter long) and the account was in cash, invest all money:
\[
s_t=\frac{(1-c)\,V_{t-1}}{P_t},\qquad \text{cash}_t=0.
\]
If $a_t=-1$ (exit to cash) and the account was fully invested, liquidate all shares:
\[
s_t=0,\qquad \text{cash}_t=(1-c)\,V_{t-1}.
\]
If $a_t=0$ (hold), or the action repeats the current regime (already all-in or all-cash), no trade occurs:
\[
s_t=s_{t-1},\qquad \text{cash}_t=\text{cash}_{t-1}.
\]
The portfolio then satisfies
\[
V_t=\text{cash}_t+s_t P_t,
\]
with costs incurred only on regime switches (buy/sell), i.e., when a trade occurs.

\textbf{RL agents.}
We integrate this environment with two standard reinforcement learning algorithms:
(i)~Deep Q-Network (DQN)~\cite{mnih2015human}, which learns a state–action value function 
$Q_\theta(s,a)$ via temporal-difference learning, and
(ii)~Proximal Policy Optimization (PPO)~\cite{schulman2017proximal}, 
which optimizes a clipped surrogate objective for the policy $\pi_\theta(a|s)$.
Both agents share the same reward structure and hyper-parameters. 

% \begin{figure}
%     \centering
%     \includegraphics[width=0.8\linewidth]{figs/system.drawio.png}
%     \caption{Model Integration}
%     \label{fig:model integration}
% \end{figure}

\subsection{Main Results}
\subsubsection{Forecasting}
Five representative task models from different model families were used to test our method. GRU and LSTM capture temporal dependencies, Transformers handle long-range dependencies with self-attention, TCN employs convolution for sequence modeling, and DLinear uses linear decomposition. We train the models on a one-day close return forecasting task with a 60-step lookback window and evaluate their performance using MSE, MAE, and the standard deviation of the loss over each timestep in the whole test range to assess the prediction robustness as shown in Table \ref{tab:main_results}.
Results show that our method achieves the best performance, significantly reducing MSE, MAE, and STD across all models. Additionally, we observe that different models respond differently to augmented data. Stronger models such as GRU, LSTM, and Transformer, which have smaller initial losses, perform well with workflows like RandAug and TrivialAug. Conversely, applying these workflows to weaker models such as DLinear and TCN may have a detrimental effect, highlighting the need for an adaptive planner to ensure a model-agnostic workflow. Furthermore, the performance improvement over AdaAug demonstrates the efficacy of the scheduler.

\begin{table}[htbp]
\centering

\begin{threeparttable}
\caption{Performance comparison of trading results when our method is integrated with RL methods.}
\label{tab:RL_result}
\centering
% \small
\setlength{\tabcolsep}{4pt}
\begin{tabular}{l|rr|rr|rr}
\toprule
\multirow{2}{*}{Method}  & \multicolumn{2}{c|}{MCD} & \multicolumn{2}{c|}{IBM} & \multicolumn{2}{c}{INTC} \\
 & \textit{TR}$\uparrow$ & \textit{SR}$\uparrow$  & \textit{TR}$\uparrow$ & \textit{SR}$\uparrow$ & \textit{TR}$\uparrow$ & \textit{SR}$\uparrow$  \\
\midrule
% Buy and Hold & 8.25  & 4.09 & 5.02 & 5.25 & 13.21 & 13.55\\
% \hline
DQN  & 4.78 & 5.06 &  13.21 & 13.55 &35.99 &16.80\\
DQN + Ours   & 17.73 & 25.74 & 13.88 & 14.80 & 33.35&21.60\\
 \hline
PPO  &     15.42 & 21.01 & -3.62   & -7.43 &34.67 &17.49 \\
PPO + Ours &   18.13 & 26.31 & -2.80 & -5.68 & 52.91 &23.45 \\
\bottomrule
\end{tabular}
\end{threeparttable}
\end{table}

\subsubsection{Transfer to Reinforcement Learning Trading}

\begin{figure}[h!]
    \centering
    \begin{subfigure}[b]{\columnwidth}
        \centering
        \includegraphics[width=\linewidth]{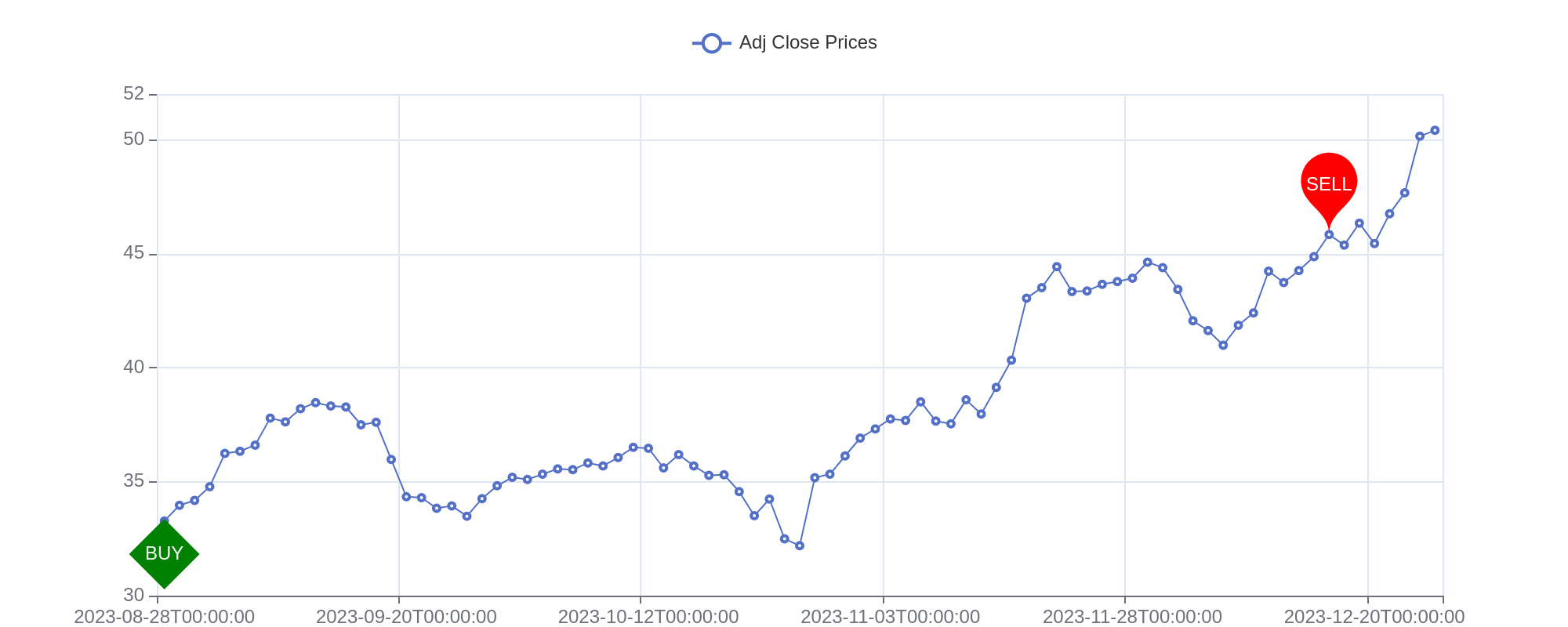}
        \caption{DQN on INTC}
        \label{fig:MDM_adjclose}
    \end{subfigure}

    \vspace{0.3cm}

    \begin{subfigure}[b]{\columnwidth}
        \centering
        \includegraphics[width=\linewidth]{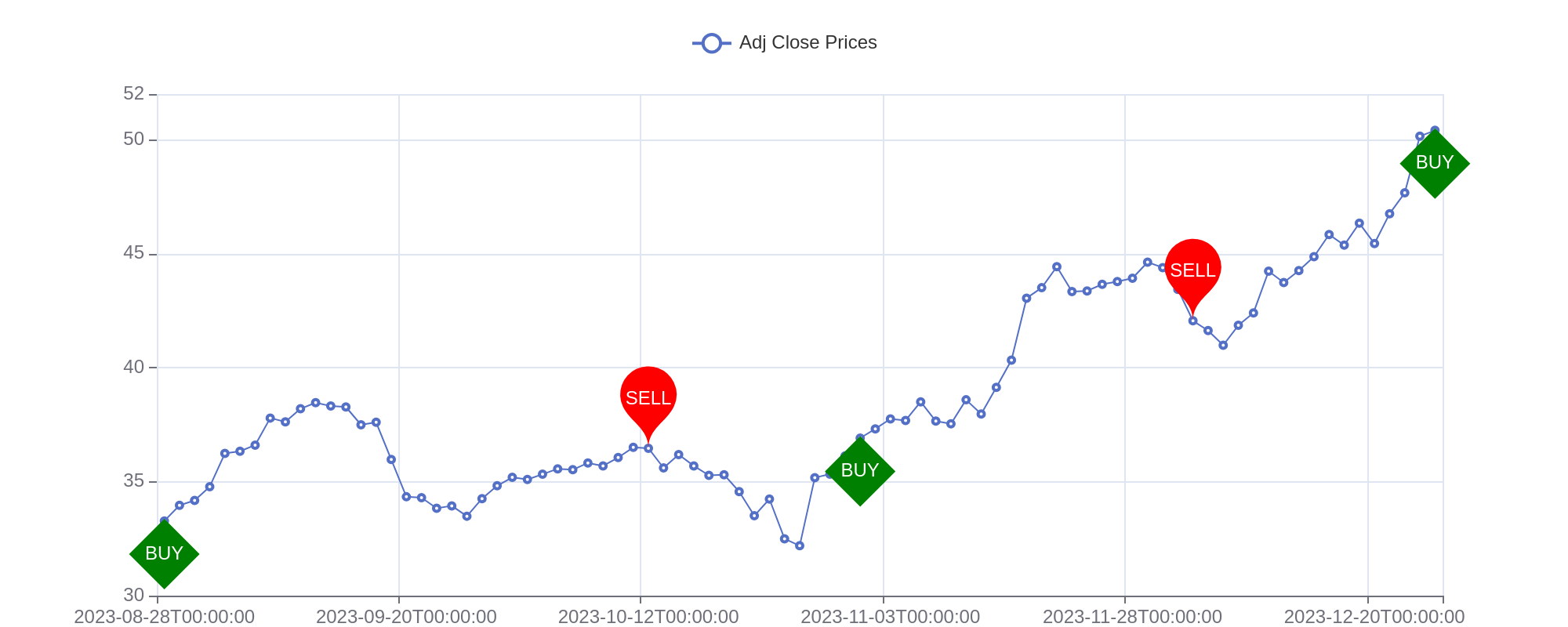}
        \caption{DQN + Ours on INTC}
        \label{fig:MDM_TSNE}
    \end{subfigure}

    \caption{Trading results where buy and sell actions are marked with green and red labels.}
    \label{fig:Trading_result}
\end{figure}

A key quantitative trading task is utilizing reinforcement learning (RL) for trading decisions \cite{qin2024earnhft, zong2024macrohft}. To evaluate the transferability of a learned policy, we conducted an experiment where the planner $g_{\phi}$, trained on a one-day return forecasting task with an LSTM task model, was applied to a single-stock trading task using DQN \cite{mnih2015human} and PPO \cite{schulman2017proximal}. To ensure fair comparisons without introducing additional data, all mix-up operations were removed. The trained planner $g_{\phi}$ was then integrated into the RL single-stock training workflow. For the scheduler, we used a simplified approach where $\alpha=0$ was set for the first $2\times10^{5}$ training steps, $\alpha=0.05$ was applied for the next $1\times10^{5}$ steps, and $\alpha=0$ was again set for the remaining steps to ensure convergence. For both DQN and PPO, the embedding dimensions were set to 128, depth at 1, initial amount at $1\times10^{4}$, transaction cost at $1\times10^{-3}$ and discount rate $\gamma$ at 0.99. For DQN, the policy learning rate is set at $2.5\times10^{-4}$, exploration fraction at 0.5, train frequency at 10, batch size at 128 and target network update frequency at 500. For PPO, the policy learning rate is set at $5\times10^{-7}$, value learning rate at $1\times10^{-6}$, generalized advantage estimation $\lambda$ at 0.95, value function coefficient at 0.5, entropy coefficient at 0.01 and target KL at 0.02. We use the \textit{Total Return} (TR) to measure profitability, and the \textit{Sharpe Ratio} (SR) to assess risk control ability. Formally, they are defined as:
\begin{equation}
\text{TR} = \frac{P_t - P_0}{P_0}
\end{equation}
where $P_t$ and $P_0$ denote the final and initial portfolio values, respectively.
\begin{equation}
\text{SR} = \frac{\mathbb{E}[\text{return}]}{\sigma(\text{return})}
\end{equation}
where $\mathbb{E}[\text{return}]$ denotes the expected return and $\sigma(\text{return})$ represents the standard deviation of returns.
As shown in Table \ref{tab:RL_result}, our method increased profit and reduced risk, demonstrating the transferability. We conducted a case study of the DQN results on INTC, where we obtained a slightly lower total return (TR) but a much higher risk-adjusted return (SR) than the baseline method, indicating superior risk control, as shown in Fig.~\ref{fig:Trading_result}. From the trading decisions, we can see that our method helps the DQN make more prudent actions, such as selling holdings before a downtrend, thus avoiding risk. This improved performance may result from encountering more diverse market scenarios during the training stage, alleviating the poor generalization caused by concept drift.

\subsection{Ablation Study}

To assess the contribution of each component in our proposed data manipulation pipeline, we perform ablation experiments on the cryptocurrency dataset by systematically disabling individual modules. Removing the \textit{multi-stock mixup} module leads to consistent increases in MSE, MAE, and STD across all forecasting models, indicating that cross-asset information is beneficial for learning more generalizable temporal dynamics. Likewise, replacing our adaptive scheduler with a fixed one—corresponding to the adapted AdaAug baseline—results in degraded performance, highlighting the importance of a learnable curriculum that adjusts augmentation intensity over time. Furthermore, disabling both the scheduler and planner (i.e., using randomly selected augmentations as in TrivialAug and RandAug) yields further performance deterioration, particularly for TCN and Transformer architectures. These findings suggest that augmentations must be dynamically and model-specifically controlled throughout training. Overall, the results confirm that the combination of mixup, adaptive scheduling, and a learned planner substantially enhances both the robustness and predictive accuracy of our forecasting framework.

\begin{figure}[htbp]
    \centering
    \begin{subfigure}{\linewidth}
        \centering
        \includegraphics[width=0.83\linewidth]{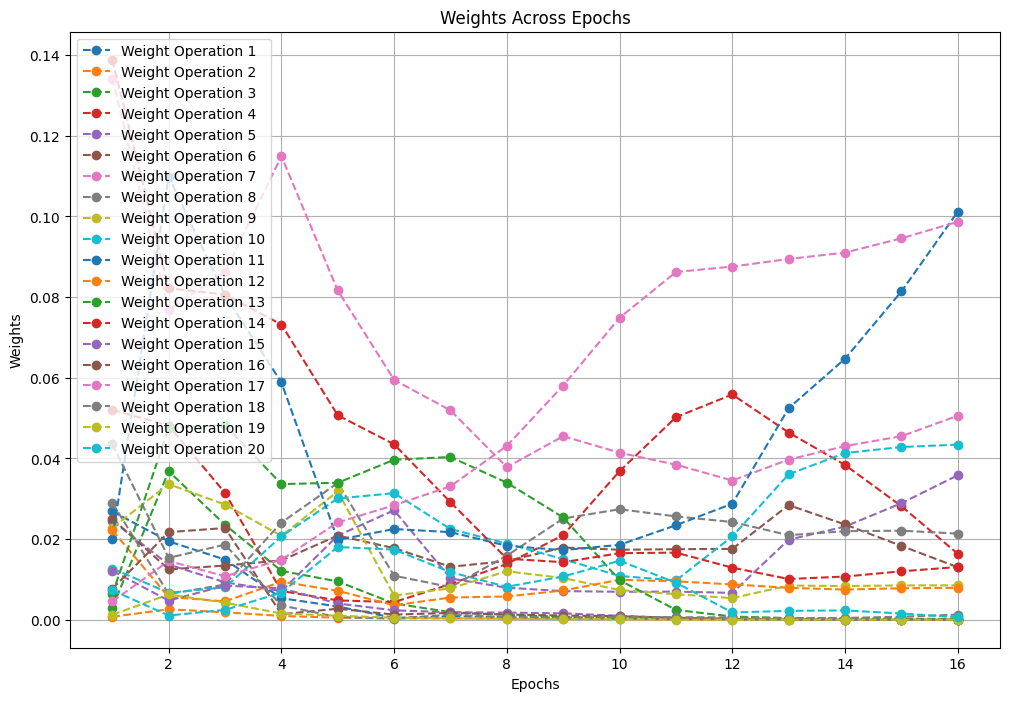}
        \caption{Transformer}
        \label{fig:weight1}
    \end{subfigure}

    \begin{subfigure}{\linewidth}
        \centering
        \includegraphics[width=0.83\linewidth]{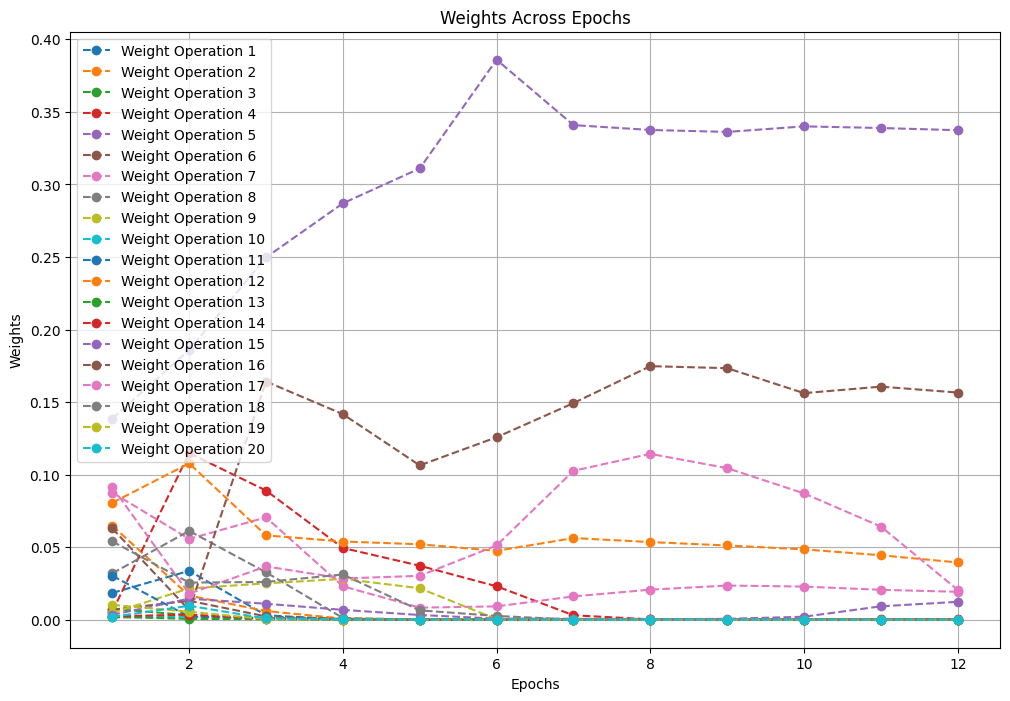}
        \caption{LSTM}
        \label{fig:weight2}
    \end{subfigure}

    \caption{Operation weights from planner for (a) Transformer and (b) LSTM.}
    \label{fig:weights}
\end{figure}

\begin{figure*}[htbp]
    \centering
    % Use tighter column specs (fewer spaces) so the row labels sit closer:
    \begin{tabular}{@{}c@{}c@{}c@{}c@{}c@{}}

        % Row of column headings:
        \multicolumn{1}{c}{}
            & \footnotesize \textbf{Amplitude Mix}
            & \footnotesize \textbf{Cut Mix}
            & \footnotesize \textbf{Linear Mix}
            & \footnotesize \textbf{\cite{demirel2024finding} Mix}
        \\[2pt]

        % ---------- Row 1: Inner Permute ----------
        % Use a smaller parbox width to bring label closer:
        \rotatebox{90}{\parbox[c]{1.8cm}{\centering \footnotesize \textbf{Inner Permute}}}
            &
            \begin{subfigure}[b]{0.14\textwidth}
                \centering
                \includegraphics[width=\textwidth]{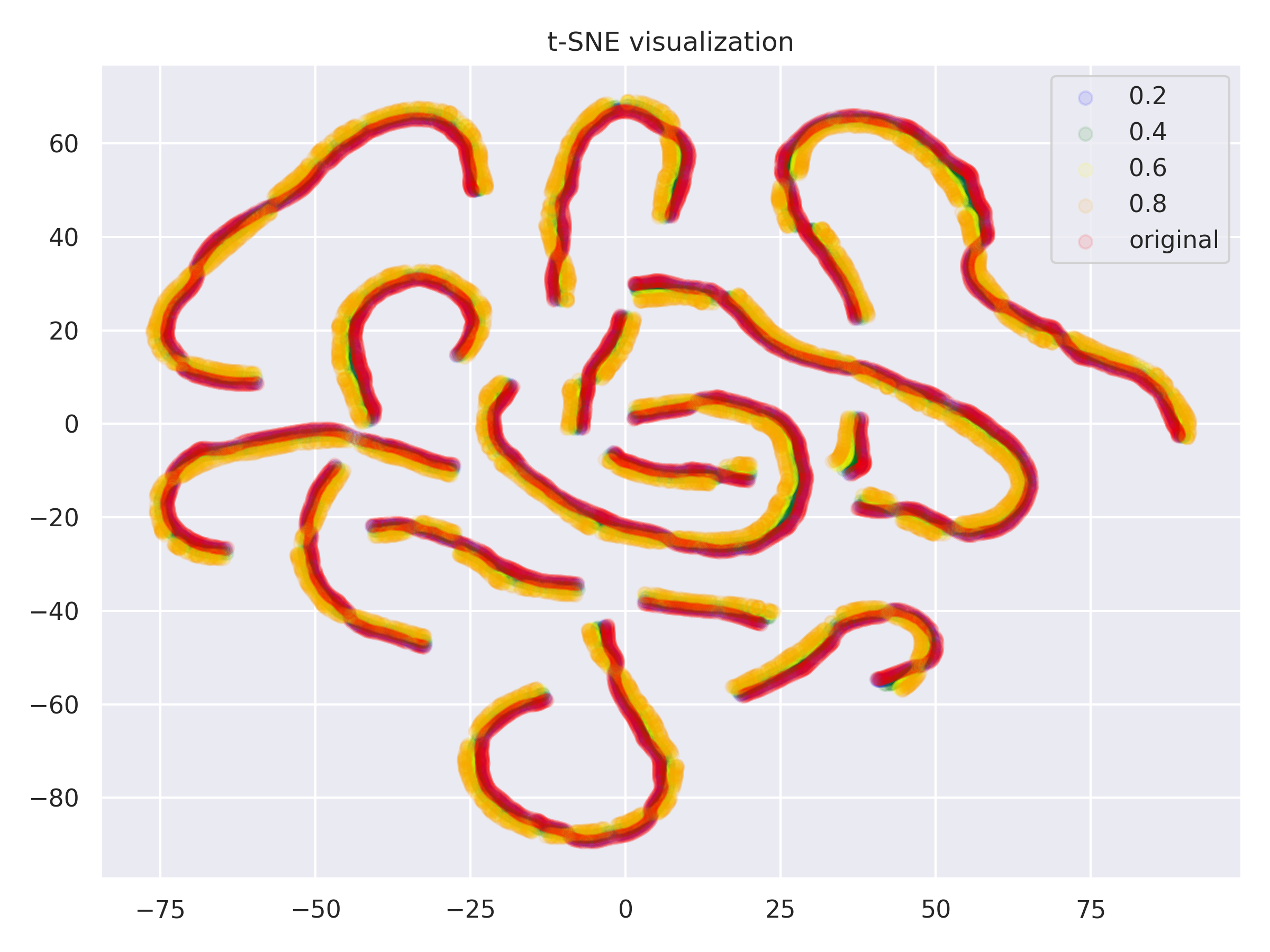}
            \end{subfigure}
            &
            \begin{subfigure}[b]{0.14\textwidth}
                \centering
                \includegraphics[width=\textwidth]{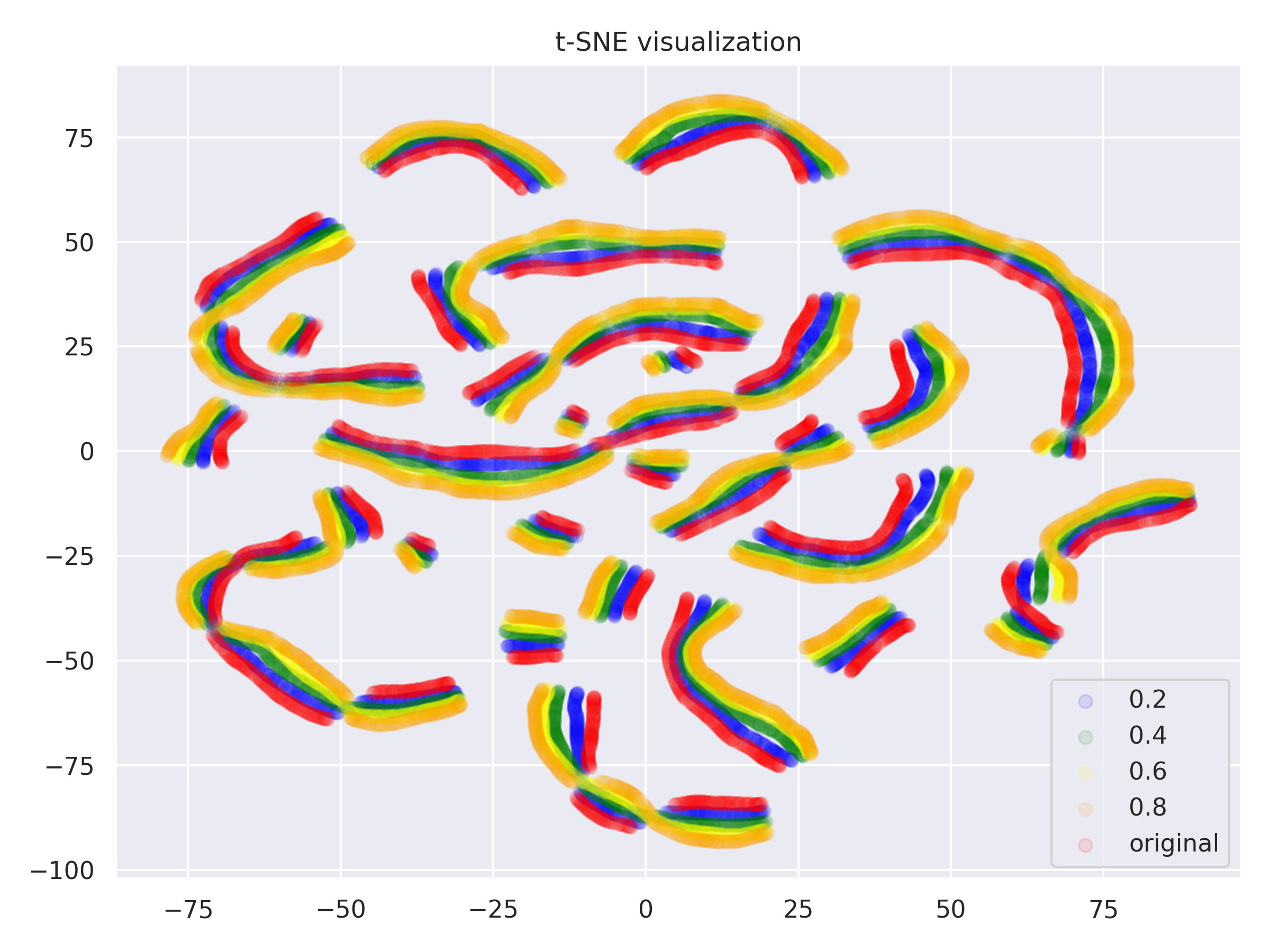}
            \end{subfigure}
            &
            \begin{subfigure}[b]{0.14\textwidth}
                \centering
                \includegraphics[width=\textwidth]{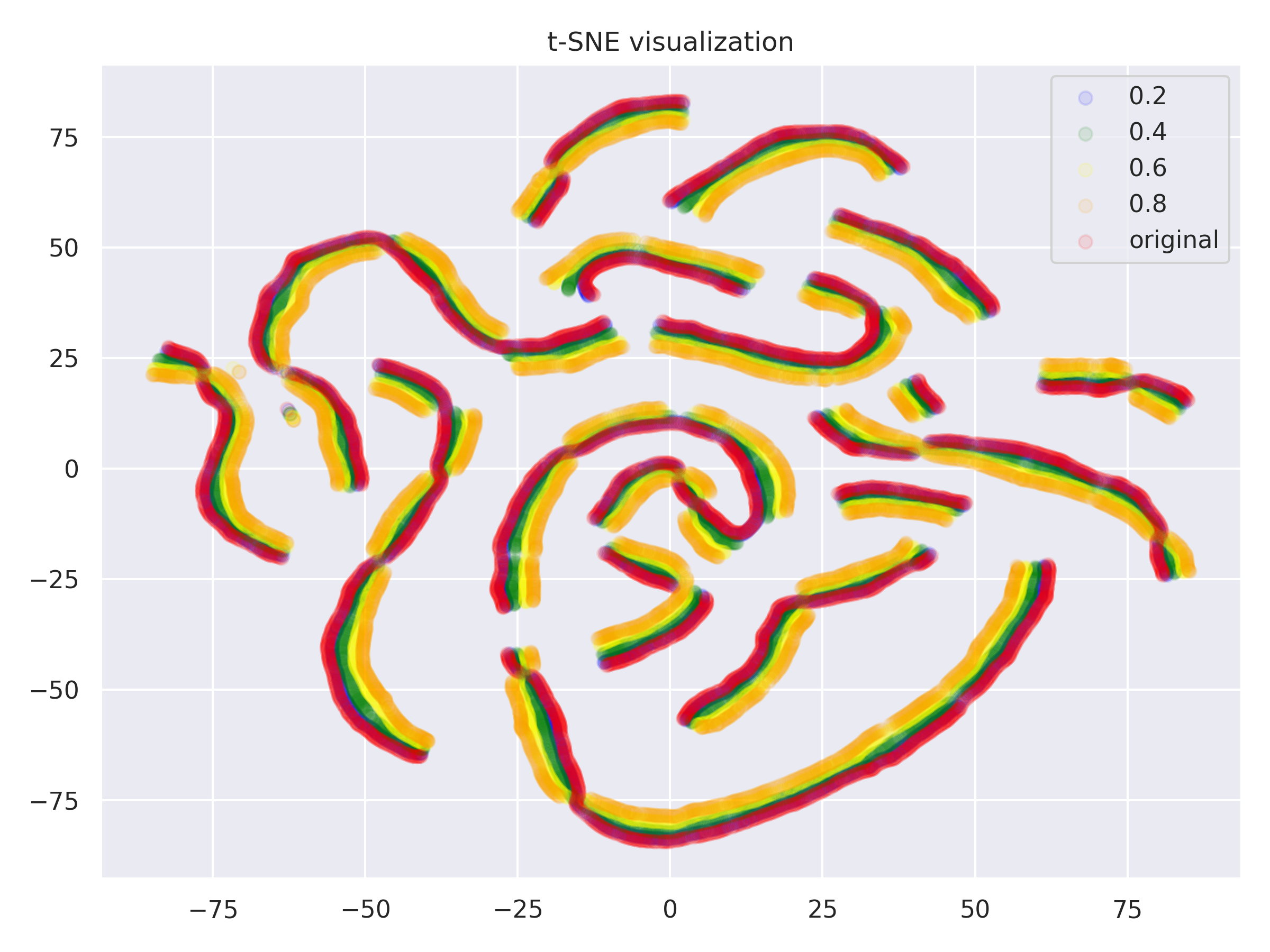}
            \end{subfigure}
            &
            \begin{subfigure}[b]{0.14\textwidth}
                \centering
                \includegraphics[width=\textwidth]{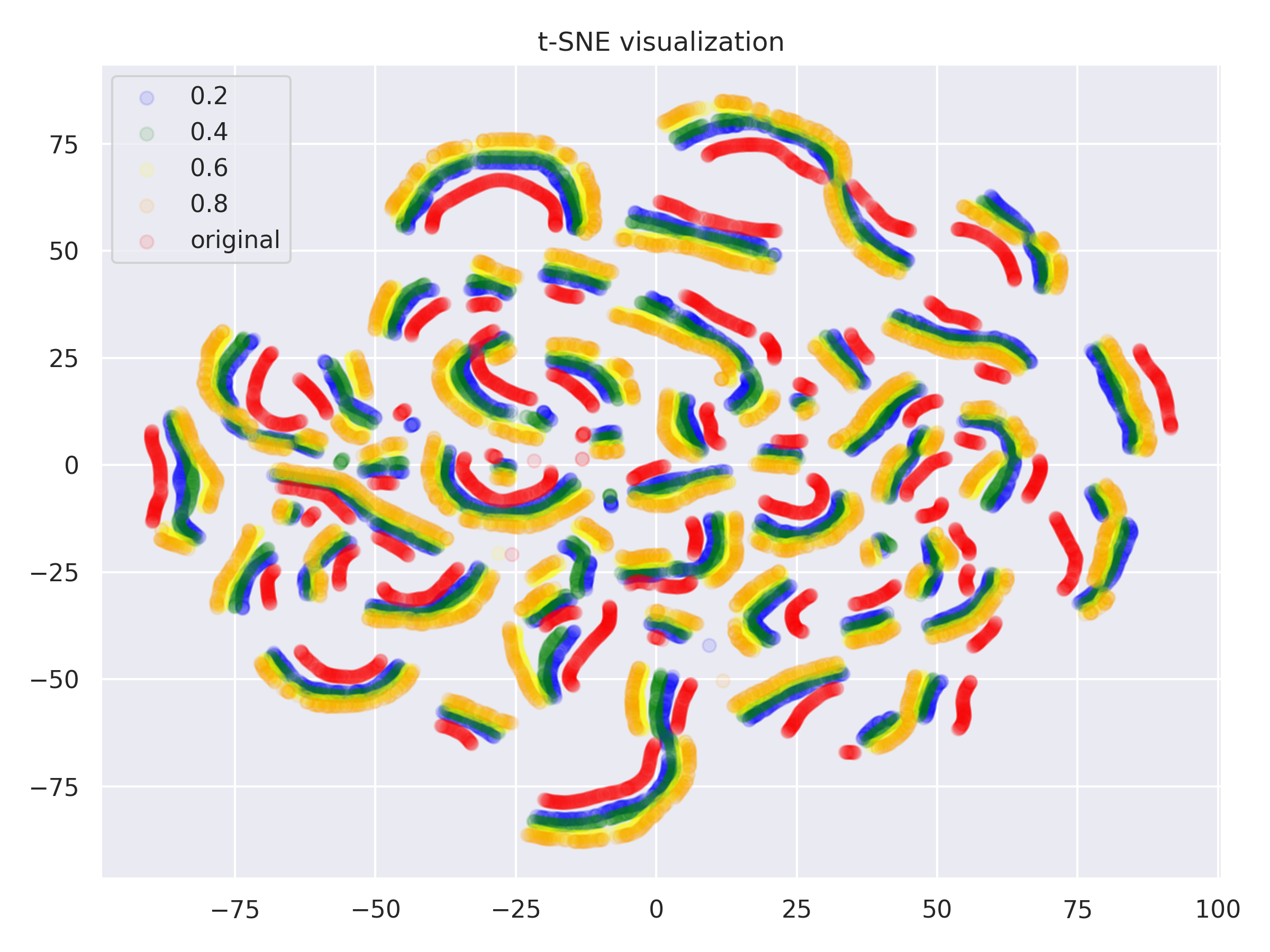}
            \end{subfigure}
        \\[0.4em]

        % ---------- Row 2: Jittering ----------
        \rotatebox{90}{\parbox[c]{1.8cm}{\centering \footnotesize \textbf{Jittering}}}
            &
            \begin{subfigure}[b]{0.14\textwidth}
                \centering
                \includegraphics[width=\textwidth]{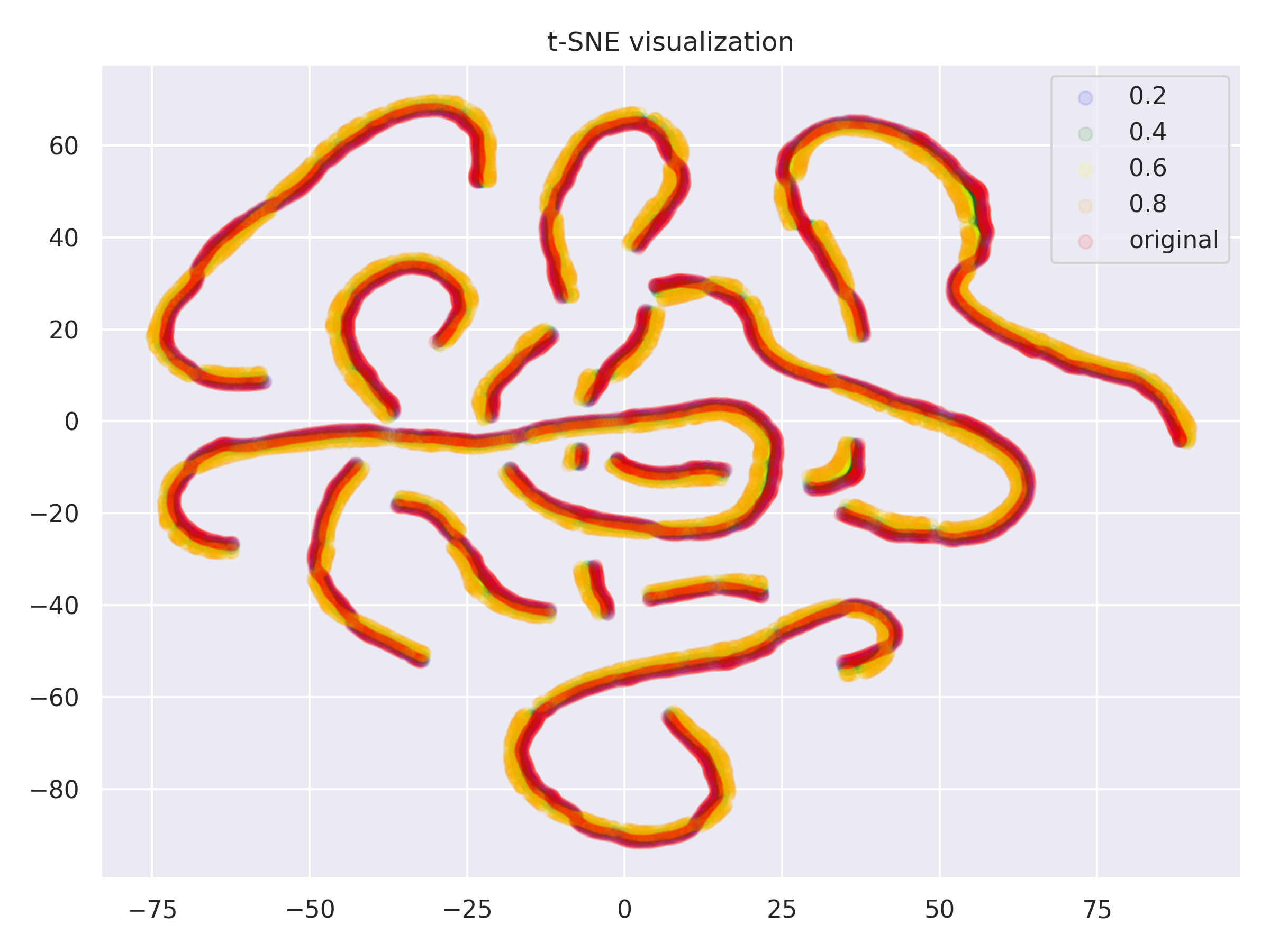}
            \end{subfigure}
            &
            \begin{subfigure}[b]{0.14\textwidth}
                \centering
                \includegraphics[width=\textwidth]{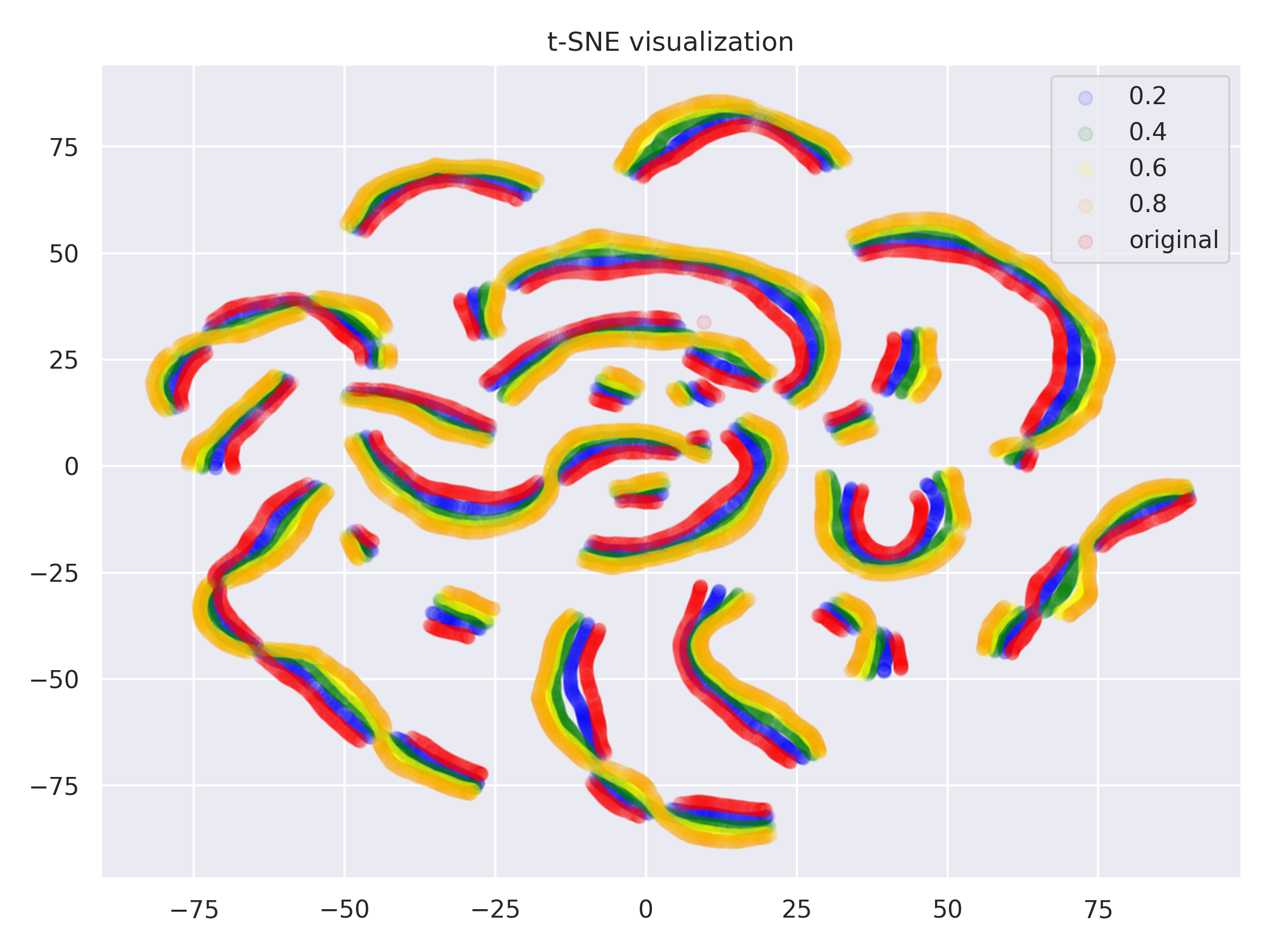}
            \end{subfigure}
            &
            \begin{subfigure}[b]{0.14\textwidth}
                \centering
                \includegraphics[width=\textwidth]{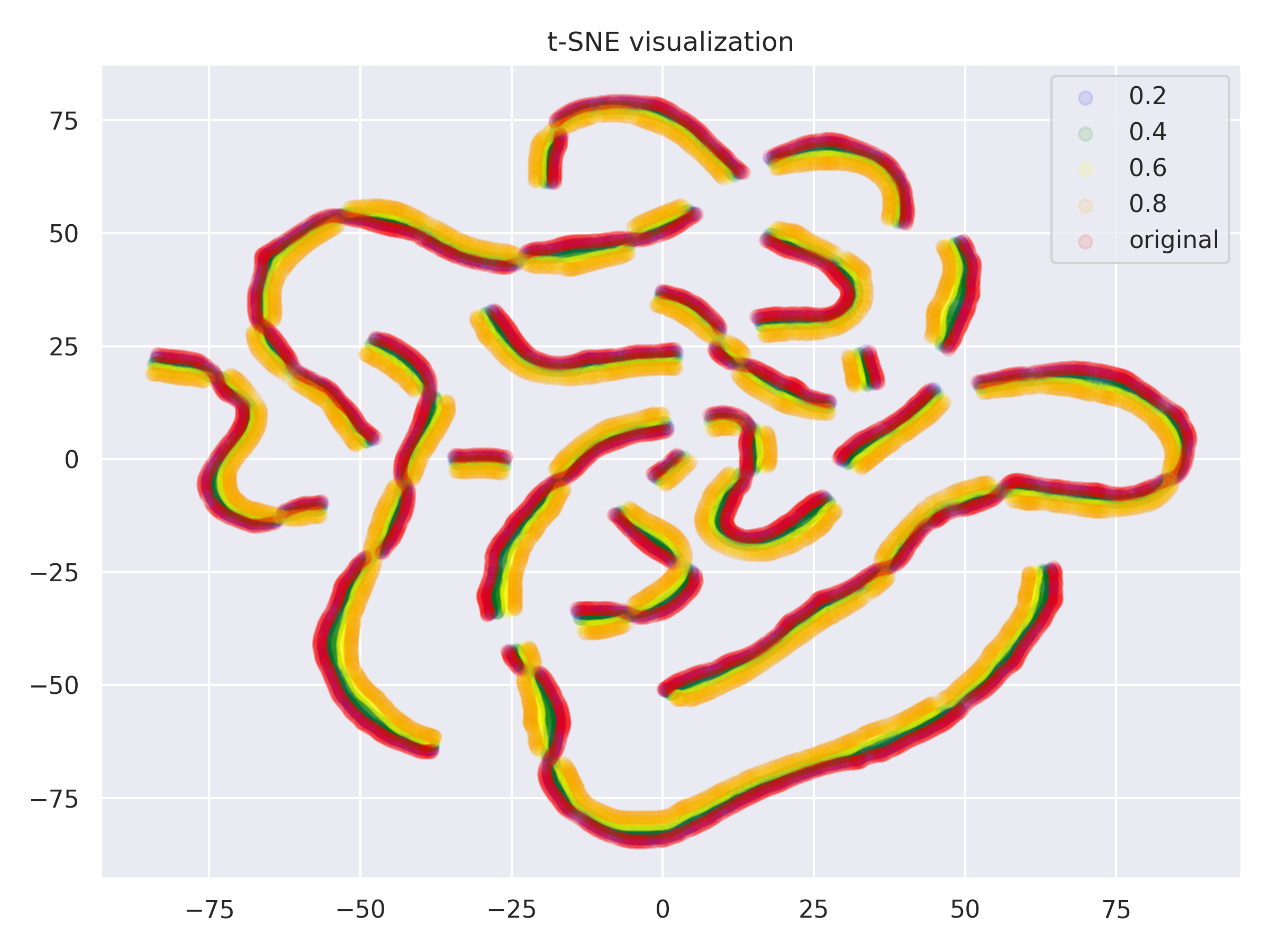}
            \end{subfigure}
            &
            \begin{subfigure}[b]{0.14\textwidth}
                \centering
                \includegraphics[width=\textwidth]{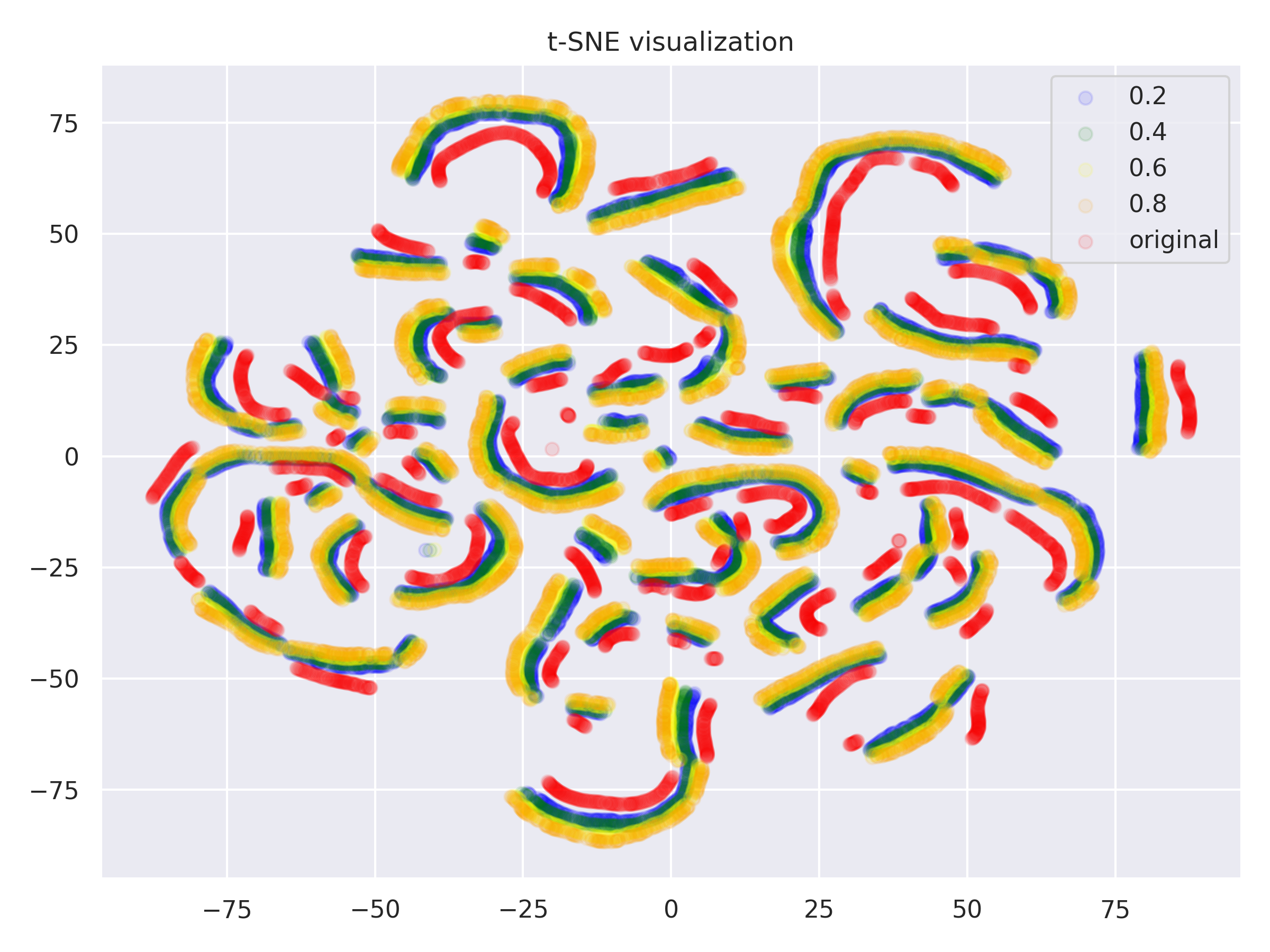}
            \end{subfigure}
        \\[0.4em]

        % ---------- Row 3: Permutation ----------
        \rotatebox{90}{\parbox[c]{1.8cm}{\centering \footnotesize \textbf{Permutation}}}
            &
            \begin{subfigure}[b]{0.14\textwidth}
                \centering
                \includegraphics[width=\textwidth]{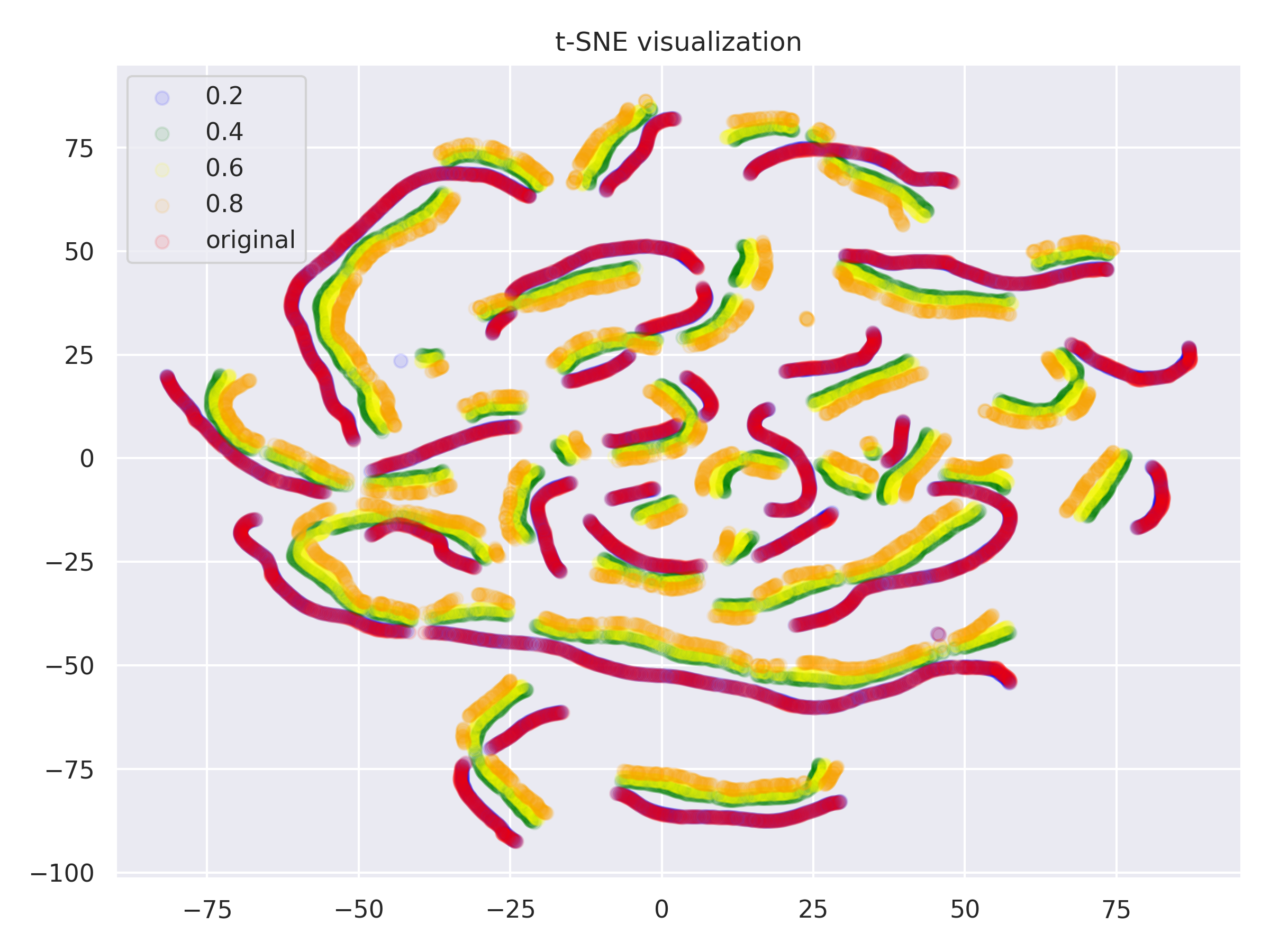}
            \end{subfigure}
            &
            \begin{subfigure}[b]{0.14\textwidth}
                \centering
                \includegraphics[width=\textwidth]{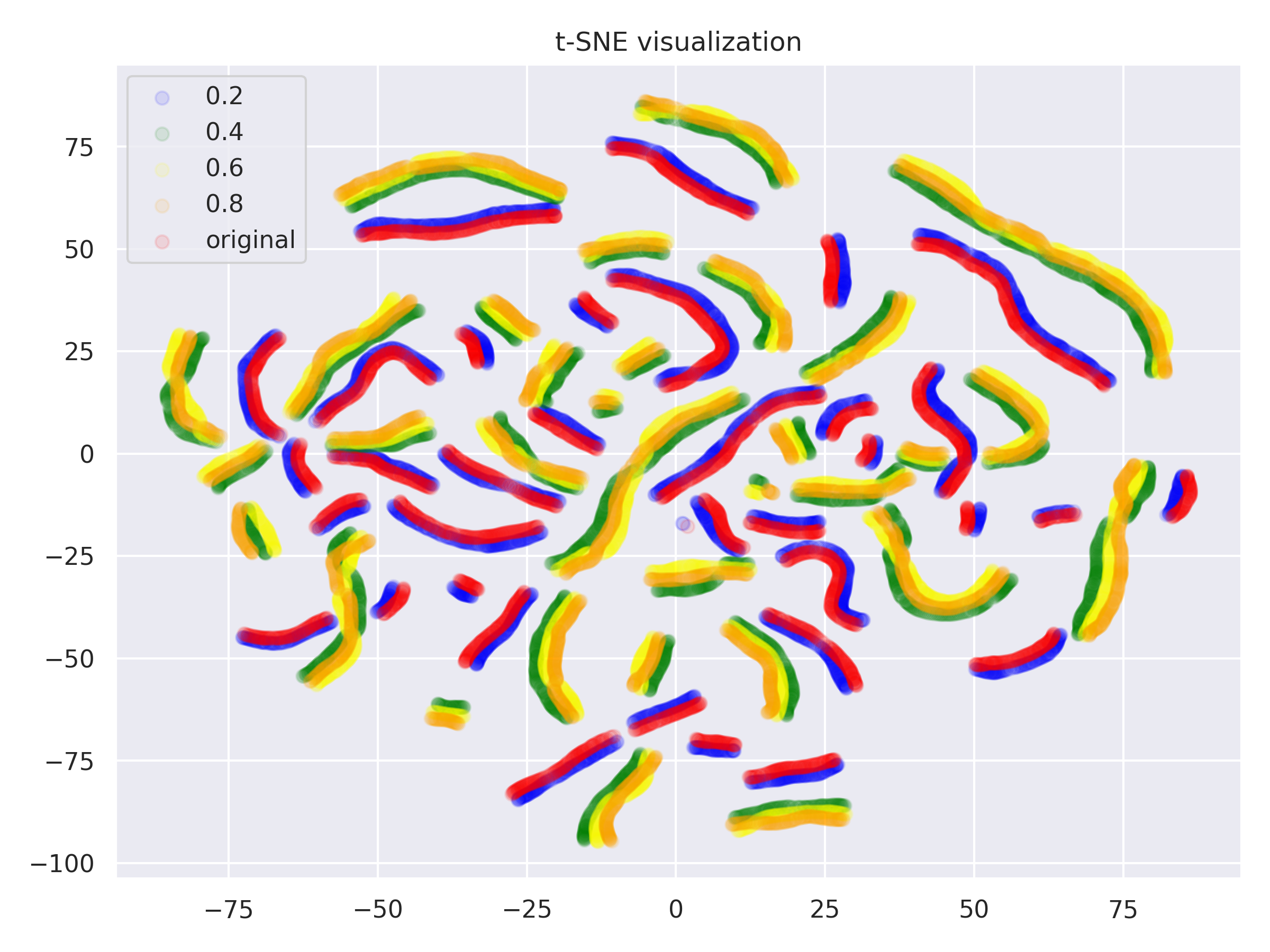}
            \end{subfigure}
            &
            \begin{subfigure}[b]{0.14\textwidth}
                \centering
                \includegraphics[width=\textwidth]{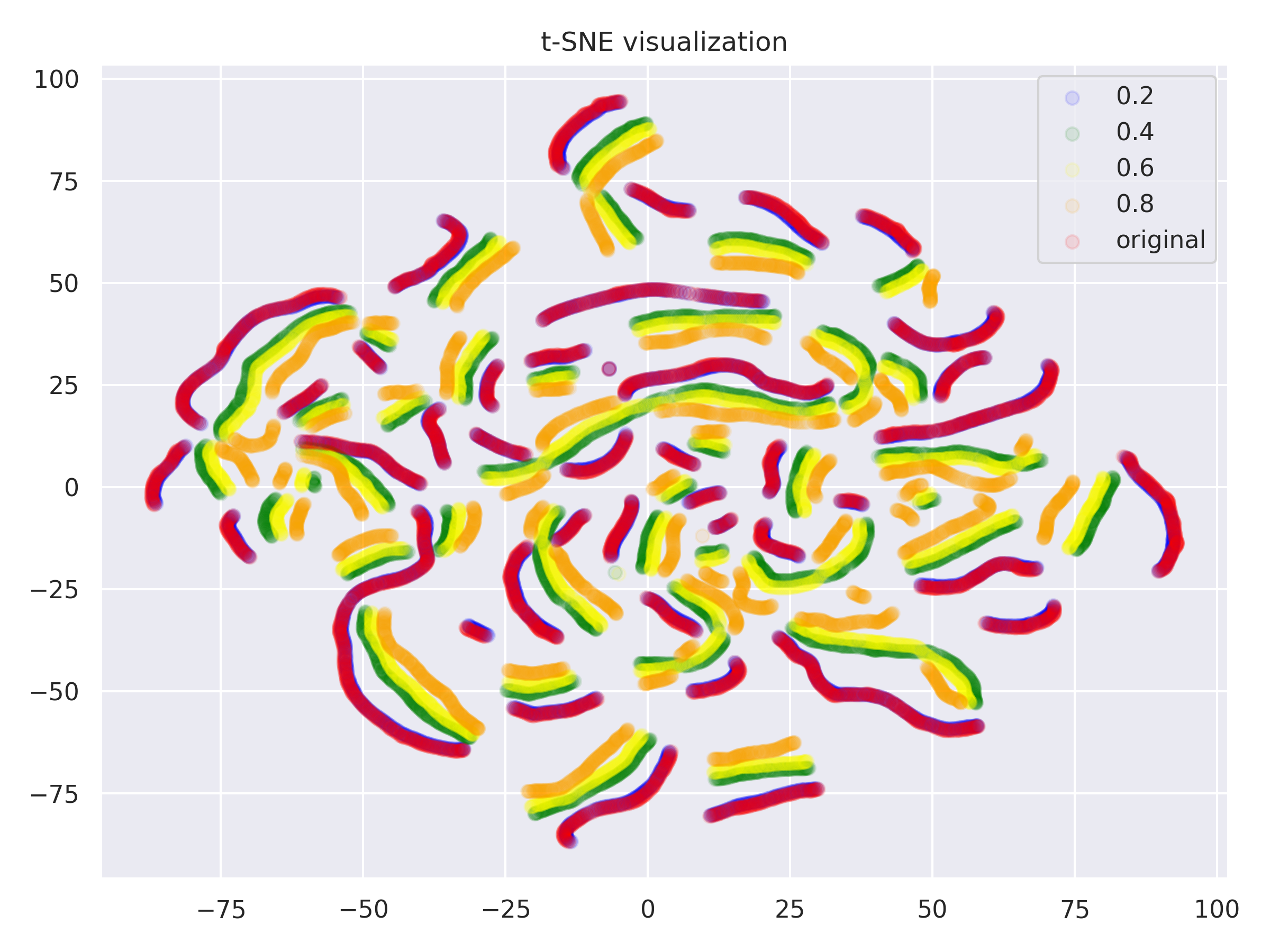}
            \end{subfigure}
            &
            \begin{subfigure}[b]{0.14\textwidth}
                \centering
                \includegraphics[width=\textwidth]{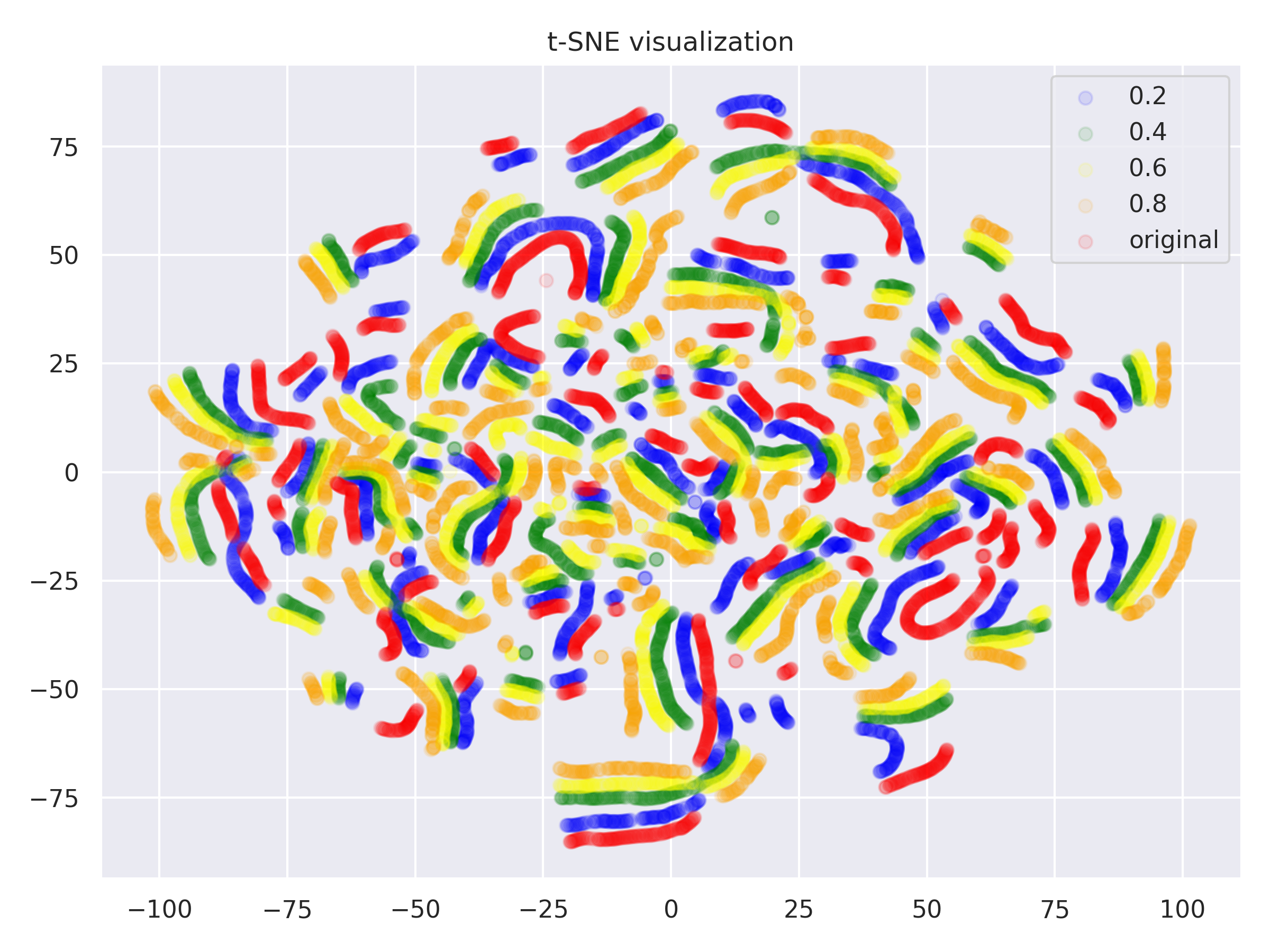}
            \end{subfigure}
        \\[0.4em]

        % ---------- Row 4: Scaling ----------
        \rotatebox{90}{\parbox[c]{1.8cm}{\centering \footnotesize \textbf{Scaling}}}
            &
            \begin{subfigure}[b]{0.14\textwidth}
                \centering
                \includegraphics[width=\textwidth]{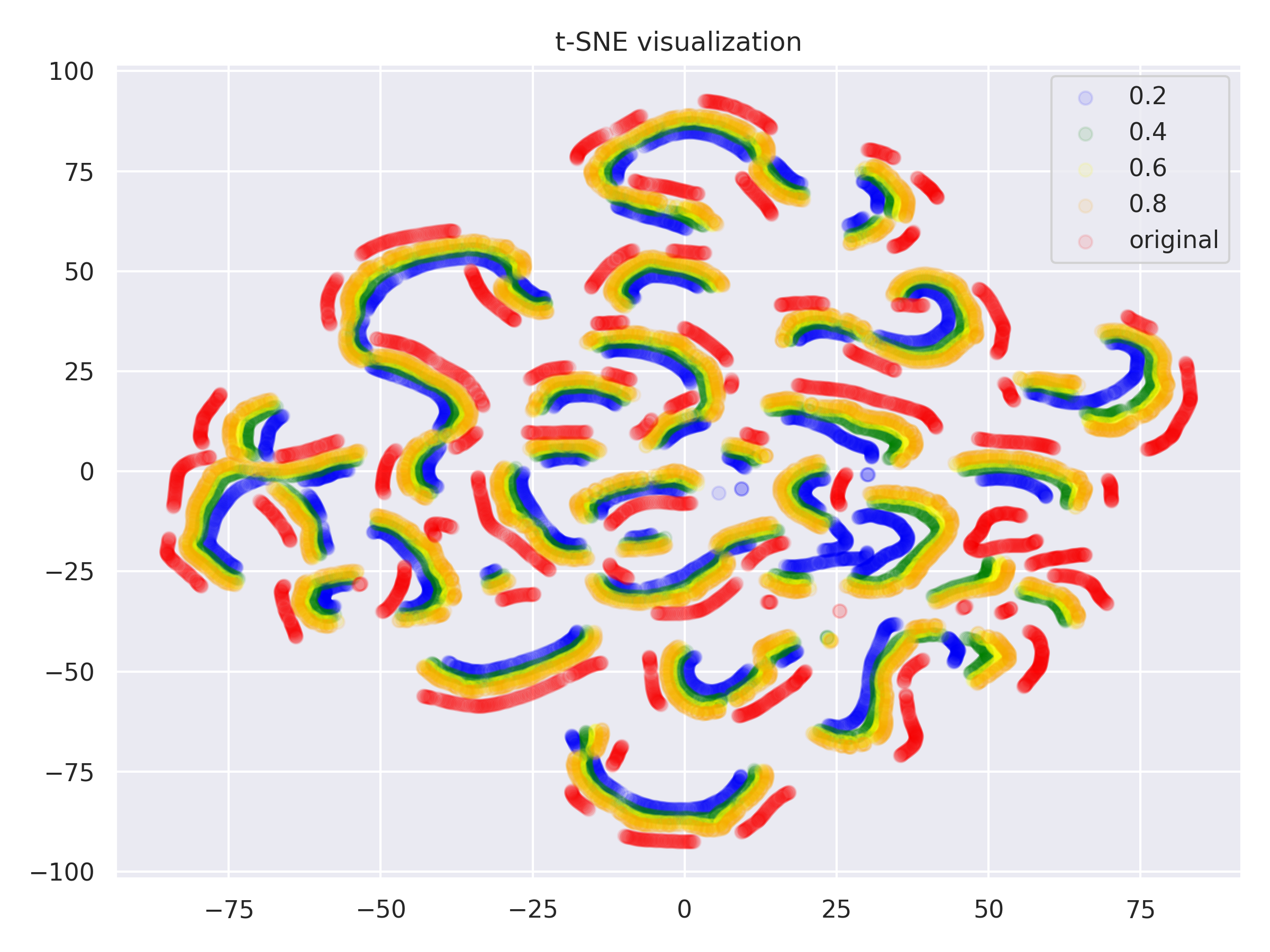}
            \end{subfigure}
            &
            \begin{subfigure}[b]{0.14\textwidth}
                \centering
                \includegraphics[width=\textwidth]{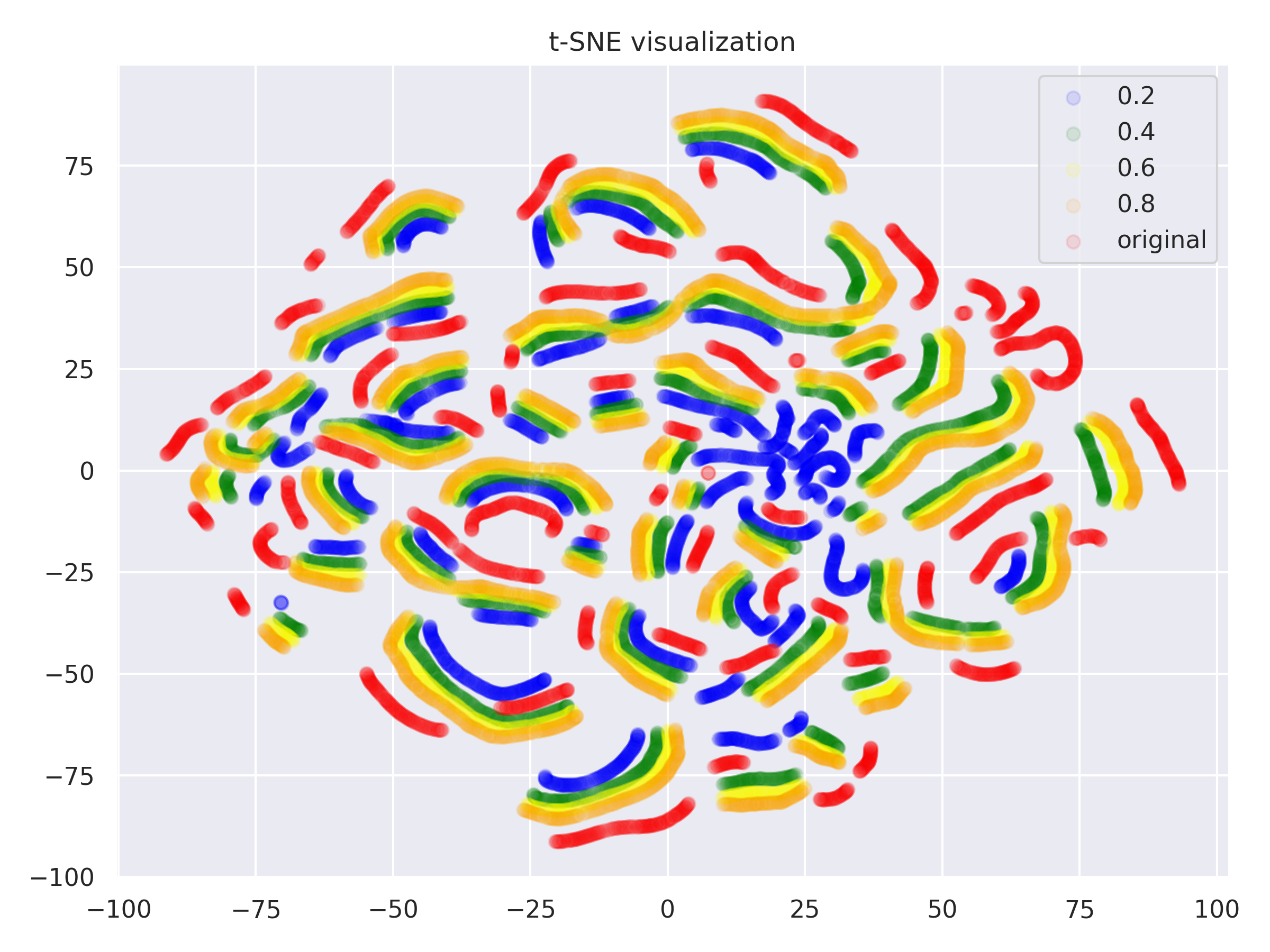}
            \end{subfigure}
            &
            \begin{subfigure}[b]{0.14\textwidth}
                \centering
                \includegraphics[width=\textwidth]{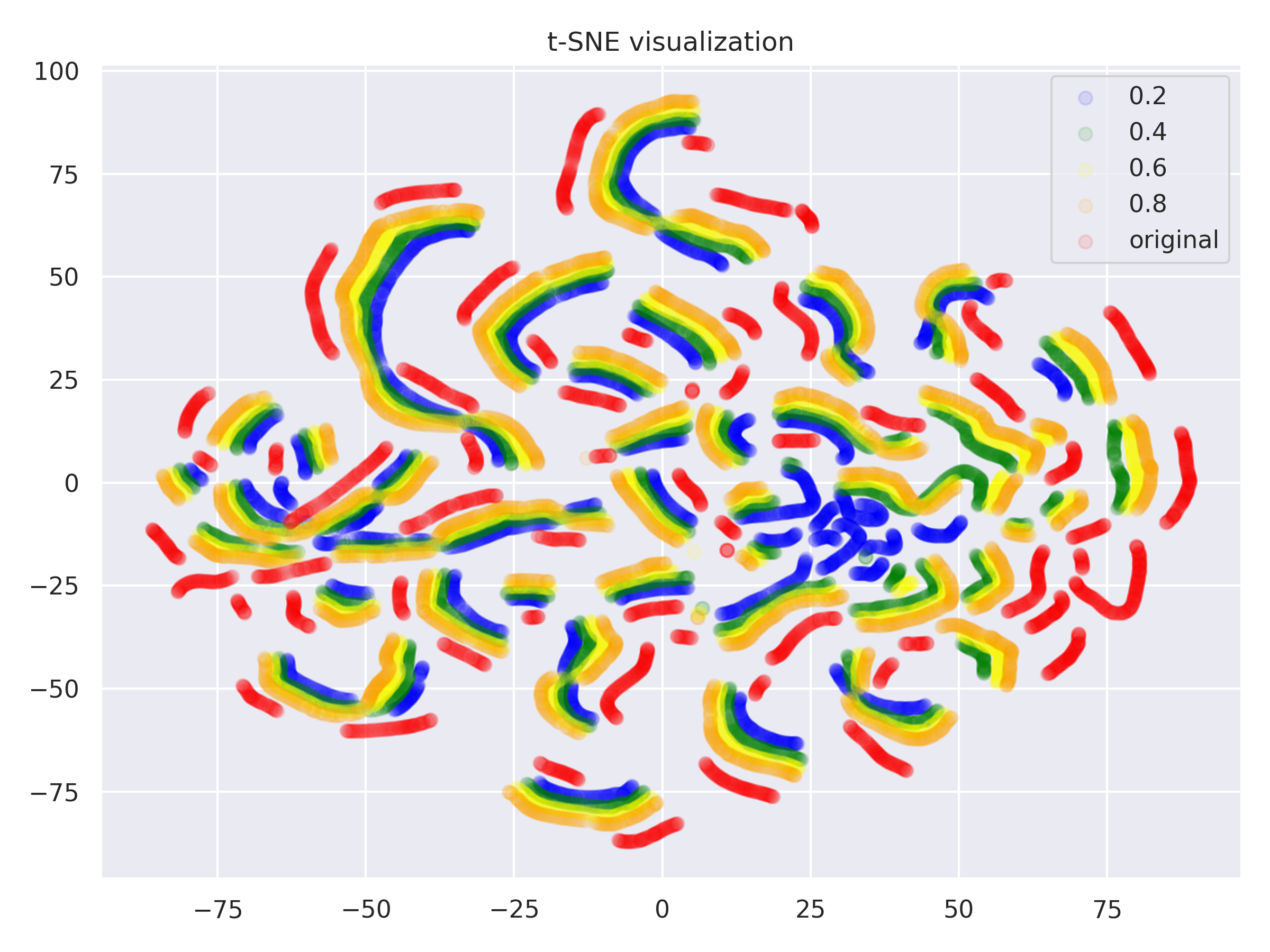}
            \end{subfigure}
            &
            \begin{subfigure}[b]{0.14\textwidth}
                \centering
                \includegraphics[width=\textwidth]{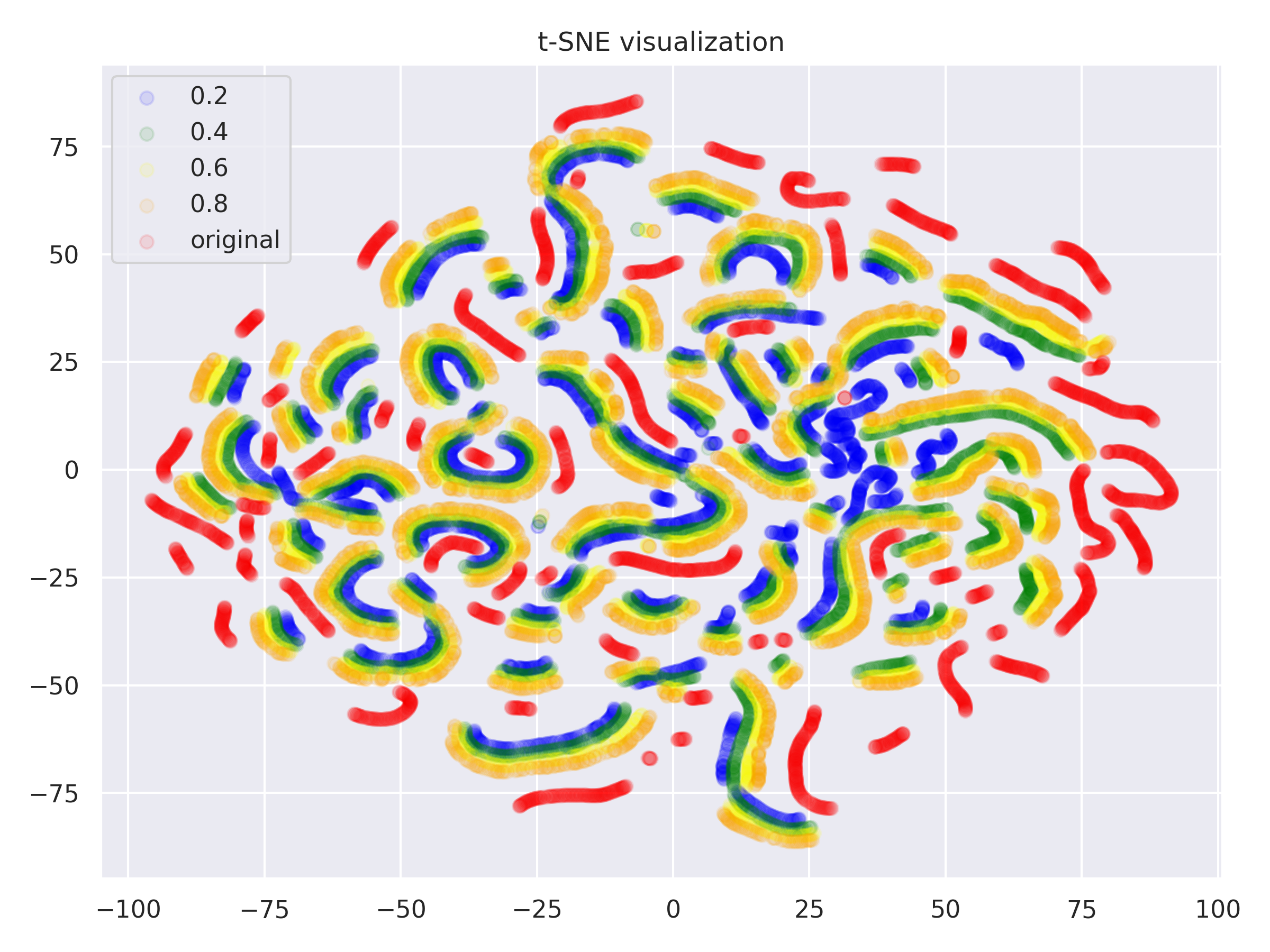}
            \end{subfigure}
        \\
    \end{tabular}
    \caption{Comparison of data distribution as the parameters vary. 
    Y-axis shows single-stock augmentations while X-axis shows multi-stock mixups. Dots of different colors represent synthetic data from varying manipulation strength $\lambda$.}
    \label{fig:augmentation-grid}
\end{figure*}
\subsection{Augmented Data Quality Evaluation}
To further ensure the safety of using augmented data, we evaluate the quality of our augmented data from a financial perspective both qualitatively and quantitatively.

%start here
%\begin{table}
% \resizebox{\linewidth}{!}{%
%\centering
%\begin{small}
%\begin{tabular}{c|c}
%\toprule
% \hline
%\textbf{Method}  & \textbf{Discriminative Score (Accuracy - 50\%)$\downarrow$} \\ 
%\midrule
%TimeGAN &               48.2\\
% \hline
%SIGCWGAN          & 48.3 \\
%RCWGAN            & 48.5 \\
%GMMN   & 49.6 \\
%CWGAN         & 48.5 \\
%RCGAN         & 48.0 \\
%\hline
%\textbf{Ours}             & \textbf{14.1}  \\

% \hline
%\bottomrule
%\end{tabular}
%\end{small}
% }
%\caption{Discriminative accuracy of post-hoc classifier.}
%\label{tab:discriminator}
%\end{table}
%end here

\begin{table}
\caption{Discriminative accuracy (Acc -- 50\%)$\downarrow$ and stylized‐facts fidelity (ACF returns, ACF absolute returns, leverage correlation) for various generative methods. A lower discriminative score and smaller stylized‐facts differences both indicate better performance.}
\centering
\begin{small}
\begin{tabular}{c|c|c|c|c}
\toprule
\textbf{Method} 
  & \textbf{Dis.} 
  & \textbf{ret} 
  & \textbf{abs ret} 
  & \textbf{Lev corr} \\
\midrule
TimeGAN    & 48.2   & 0.0231  & 0.00987 & 0.0263  \\
SIGCWGAN   & 48.3   & 0.0625  & 0.0113  & 0.0678  \\
RCWGAN     & 48.5   & 0.0278  & 0.0142  & 0.00776 \\
GMMN       & 49.6   & 0.0280  & 0.0555  & 0.146   \\
CWGAN      & 48.5   & 0.0118  & 0.03177 & 0.0309  \\
RCGAN      & 48.0   & 0.0210  & 0.0154  & 0.0251  \\
Diffusion-TS & 42.4   & 0.0361  & 0.026  &  0.022 \\
\textbf{Ours}
           & \textbf{14.1}
           & \textbf{0.000133}
           & \textbf{0.000478}
           & \textbf{0.00224} \\
\bottomrule
\end{tabular}
\end{small}
\label{tab:merged_results}
\end{table}

%  \begin{table}[htbp]
% % \resizebox{\linewidth}{!}{%
% \centering
% \begin{small}
% \begin{tabular}{c|c|c|c }
% \toprule
% % \hline
% % Original Data& -0.0274 &-0.0285 & 0.000370 \\
% % \midrule
% \textbf{Method}  & \textbf{ACF returns} & \textbf{ACF abs returns} &\textbf{Leverage corr}\\ 
% \midrule
% TimeGAN & 0.0231 &0.00987& 0.0263  \\
% % \hline
% SIGCWGAN  & 0.0625& 0.0113 & 0.0678 \\
% RCWGAN & 0.0278& 0.0142 & 0.00776 \\
% GMMN   & 0.0280 & 0.0555 & 0.146\\
% CWGAN   & 0.0118 & 0.03177& 0.0309 \\
% RCGAN   &0.0210 &0.0154 & 0.0251 \\
% \hline
% \textbf{Ours} & \textbf{0.000133} & \textbf{0.000478} & \textbf{0.00224}\\

% % \hline
% \bottomrule
% \end{tabular}
% \end{small}
% % }
% \caption{Absolute difference of market stylized facts with the original data. A smaller difference indicates greater fidelity.}
% \label{tab:stylised_facts}
% \end{table}

\begin{figure}[h!]
    \centering
    \begin{subfigure}{0.49\linewidth}
        \centering
        \includegraphics[width=\linewidth]{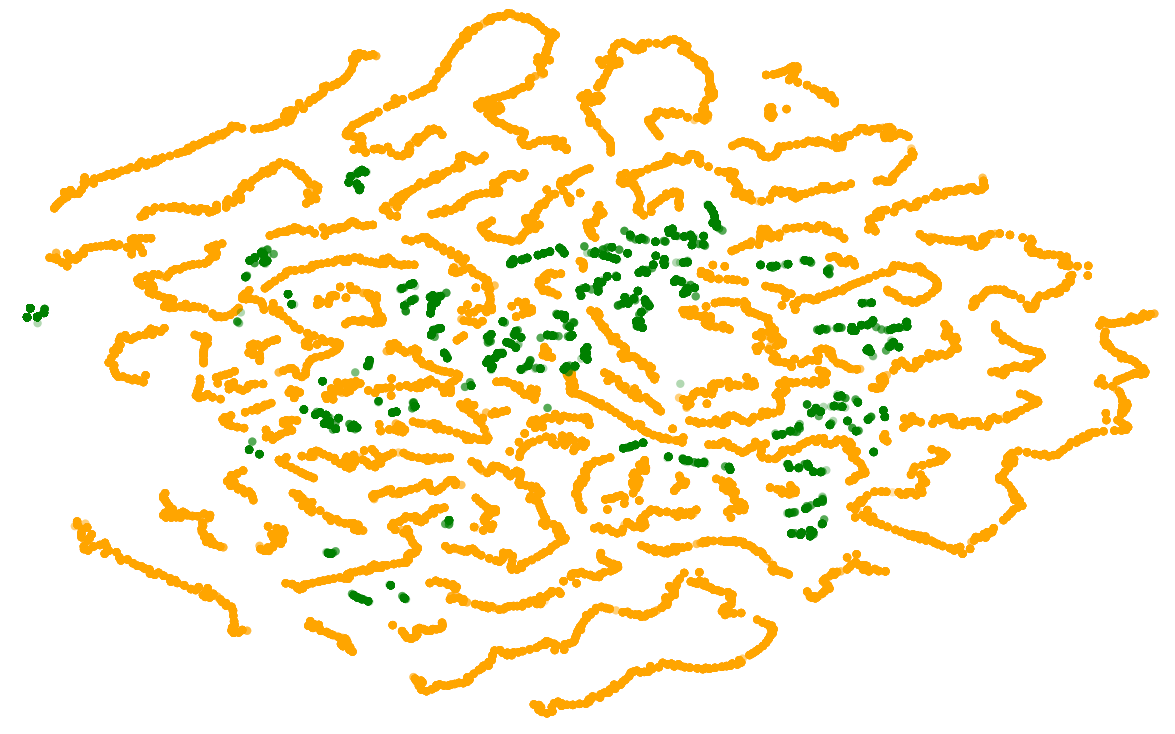}
        \caption{t-SNE of original data}
        \label{fig:MDM_adjclose}
    \end{subfigure}
    \hfill
    \begin{subfigure}{0.49\linewidth}
        \centering
        \includegraphics[width=\linewidth]{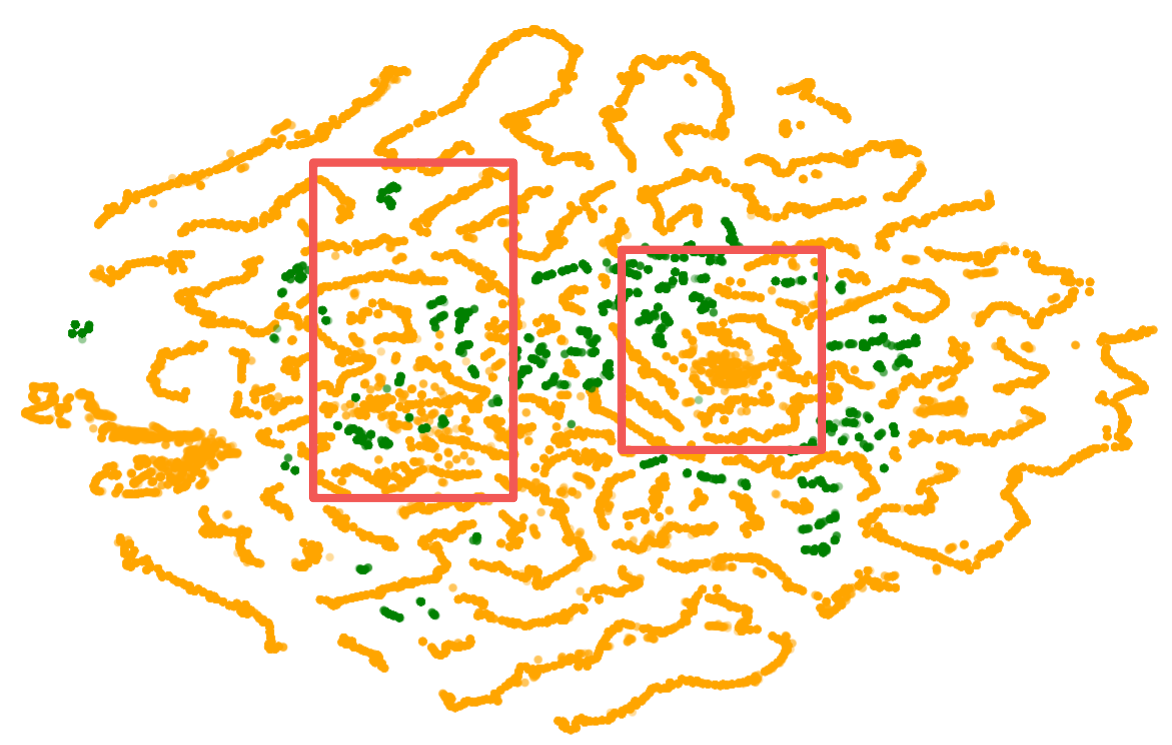}
        \caption{t-SNE of augmented data}
        \label{fig:MDM_TSNE}
    \end{subfigure}
    % \caption{Sample t-SNE plots comparing original data and augmented data from our pipeline. The red box highlights that the in-distribution data is closer to the out-distribution data.}
\caption{
t-SNE plots comparing original and augmented data.
The red box highlights that the augmented \textcolor{orange}{training data (orange)} 
becomes distributionally closer to \textcolor{green!60!black}{test data (green)}.
}
    \label{fig:tsne_compare}
\end{figure}

% concept drift TSNE
\paragraph{Addressing Concept Drift} To assess how our augmented data mitigates concept drift, we compare the training and test distributions in Fig.~\ref{fig:tsne_compare}. In (a), the original training set clusters further from the test set, indicating potential drift. In contrast, (b) demonstrates that the augmented training samples cluster more closely to the test set, highlighting the ability of our pipeline to address concept drift.

\paragraph{Variable Controllability} To visualize how our data manipulation parameters affect augmentation, we present t-SNE plots in Fig.~\ref{fig:augmentation-grid}, showing that the module is tunable and parameter changes lead to gradual, predictable shifts in the synthetic data distribution, providing diverse and controlled augmented data.

\paragraph{Downstream Usability}
\label{section: Downstream Usability}

We assess the suitability of the operations for financial time series with a general experiment. An LSTM was used to forecast the close returns of stocks in DJI index and we use the classification accuracy of the return direction as labels. The operations were applied to the entire training set respectively to determine if the augmented data was useful. The results in Table \ref{tab:preliminary_results} indicate that all selected operations improved the forecasting accuracy over the baseline of using original historical data.

% Additionally, preliminary experiments in Table \ref{tab:preliminary_results} assess the operations on financial time series, showing that all operations improve forecasting accuracy.

% varibale TSNE
%\paragraph{Variable Controllability} 

%To examine how the parameters of our data manipulation module influence the augmented data, we generate t-SNE plots in Fig.~\ref{fig:augmentation-grid}. This illustrates how varying these parameters results in incremental shifts in the synthetic data distribution, providing diverse and controlled augmented data.
% indicate that we can control the augmetated data with varibale, geting diverse data

\paragraph{Discriminative Score}
Following TimeGAN, we evaluate the fidelity of augmented data using post-hoc RNN classifiers to distinguish real from augmented data. An accuracy of $50\%$ indicates indistinguishability, and fidelity is measured as the classifier's accuracy minus $50\%$. Although our pipeline is not designed to be a generative model, we include a comparison with deep generative models. This comparison provides additional insight into the closeness of our augmented data to the real distribution. As observed in Table \ref{tab:merged_results}, our method achieves the lowest discriminative score, showing the highest financial fidelity.

% \begin{table}
% % \resizebox{\linewidth}{!}{%
% \centering
% \begin{small}
% \begin{tabular}{c|c}
% \toprule
% % \hline
% \textbf{Operation}  & \textbf{Accuracy (\%)} \\ 
% \midrule
% None &               50.72\\
% \hline
% Jittering           & 52.06 \\
% Scaling             & 51.54 \\
% Magnitude Warping   & 51.75 \\
% Permutation         & 51.95 \\
% STL Augment         & 51.75 \\
% \hline
% Cut Mix             & 52.37  \\
% Linear Mix          & 51.85  \\
% Amplitude Mix       & 52.16  \\
% \cite{demirel2024finding} Mix        & 51.65  \\
% % \hline
% \bottomrule
% \end{tabular}
% \end{small}
% % }
% \caption{Accuracy after applying operations in training.}
% \label{tab:preliminary_results}
% \end{table}

\paragraph{Market Stylized Facts}
% We compare the market-stylized facts of augmented data from our pipeline with those generated by various deep generative models to assess the consistency and fidelity of the augmented data in relation to the original data. 
Beyond classification-based metrics, it is crucial to confirm that the augmented data adheres to well-established stylized facts known to characterize real financial markets.Concretely, we compute three key statistical properties commonly observed in financial markets: the autocorrelation of returns, the autocorrelation of absolute returns, and the leverage effect. They are defined as follows:

\begin{equation}
\rho_r(k) = \frac{\text{Cov}(r_t, r_{t-k})}{\text{Var}(r_t)}
\end{equation}

\begin{equation}
\rho_{|r|}(k) = \frac{\text{Cov}(|r_t|, |r_{t-k}|)}{\text{Var}(|r_t|)}
\end{equation}

\begin{equation}
\rho_{r,\sigma}(k) = \text{Corr}(r_t, \sigma_{t+k})
\end{equation}

where $\rho_r(k)$ and $\rho_{|r|}(k)$ denote the lag-$k$ autocorrelation of returns and absolute returns, respectively, and $\rho_{r,\sigma}(k)$ represents the leverage effect. The autocorrelation of returns helps evaluate market efficiency, the autocorrelation of absolute returns is central to risk modeling, and the leverage effect is crucial for designing volatility models and assessing risk.

We evaluate how faithfully the augmented series captures core market dynamics. As shown in Table~\ref{tab:merged_results}, our augmented data exhibits stylized facts that most closely align with those found in true financial data, underscoring the practical relevance of our augmentation pipeline for financial tasks.

\begin{table}
\caption{Accuracy after applying operations in training.}
% \resizebox{\linewidth}{!}{%
\centering
\begin{small}
\begin{tabular}{c|c}
\toprule
% \hline
\textbf{Operation}  & \textbf{Accuracy (\%)} \\ 
\midrule
None &               50.72\\
\hline
Jittering           & 52.06 \\
Scaling             & 51.54 \\
Magnitude Warping   & 51.75 \\
Permutation         & 51.95 \\
STL Augment         & 51.75 \\
\hline
Cut Mix             & 52.37  \\
Linear Mix          & 51.85  \\
Amplitude Mix       & 52.16  \\
\cite{demirel2024finding} Mix        & 51.65  \\
% \hline
\bottomrule
\end{tabular}
\end{small}
% }
\label{tab:preliminary_results}
\end{table}

\subsection{Additional Analyses}

\paragraph{Operation Weights}

We visualize the weights of the probabilistic operations $p$ in Fig.~\ref{fig:weight1} and Fig.~\ref{fig:weight2} with our provenance aware replay. As observed, $p$ evolves as the task model undergoes training. The weights of these operations also vary significantly between different models, indicating that a planner is essential for a model-agnostic adaptive policy.

\begin{figure}[htbp]
    \centering
    \includegraphics[width=0.95\linewidth]{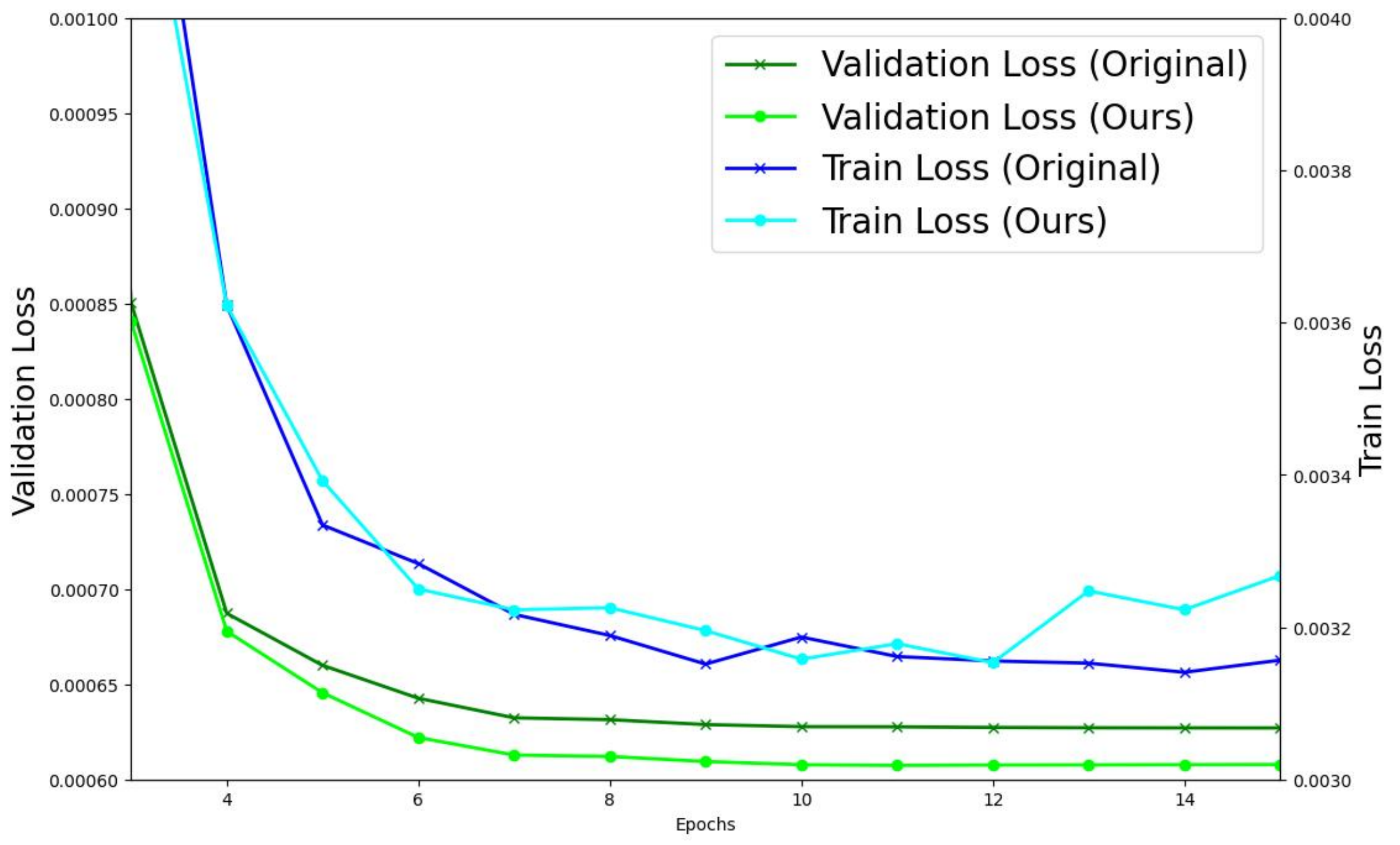}
    \caption{The training and validation loss curve w/wo our workflow applied.}
    \label{fig:training_loss}
\end{figure}

\paragraph{Learning Curve}
%  validation loss plot

We conducted a qualitative analysis to understand why our method enhances performance, as shown in Fig.~\ref{fig:training_loss}. Compared to the original workflow, our approach results in a lower validation loss, indicating successful generalization and effectively addressing potential overfitting caused by concept drift.

% weight of the operations

% \begin{figure}
%     \centering
%     \includegraphics[width=0.8\linewidth]{figs/weights/transformer.png}
%     \caption{Weights from planner for Transformer}
%     \label{fig:weight1}
% \end{figure}

% \begin{figure}
%     \centering
%     \includegraphics[width=0.8\linewidth]{figs/weights/LSTM_weights.png}
%     \caption{Weights from planner for LSTM}
%     \label{fig:weight2}
% \end{figure}

% \subsection{Operation Weights}

% We visualize the weights of the probabilistic operations $p$ in Fig.~\ref{fig:weight1} and Fig.~\ref{fig:weight2}. As observed, $p$ evolves as the task model undergoes training. The weights of these operations also vary significantly between different models, indicating that a planner is essential for a model-agnostic adaptive policy.

%put in main paper

\section{Conclusion}

In this paper, we introduced a novel adaptive dataflow system designed to bridge the gap between training and real-world performance in quantitative finance. To the best of our knowledge, this is the first workflow of its kind applied to quantitative finance tasks. The framework integrates a parameterized data manipulation module with a learning-guided planner–scheduler, forming a feedback loop that dynamically regulates manipulation strength and proportion of data to be manipulated as the model evolves.
This design allows the data pipeline to self-adjust to distributional drift, ensuring consistent data quality and realistic synthesis throughout the learning process.
Experiments on forecasting and trading tasks demonstrate that the system improves robustness and generalization across models and markets.

% \section{AI-Generated Content Acknowledgment}
% ChatGPT was lightly used across all sections of this paper to assist with language polishing, style consistency, and LaTeX formatting.

\bibliographystyle{IEEEtran}
\bibliography{sample-base}

\end{document}